\documentclass{article}

\usepackage[preprint]{robotlearn} 

\usepackage[utf8]{inputenc}
\usepackage[T1]{fontenc}
\usepackage{url}
\usepackage{booktabs}
\usepackage{amsfonts}
\usepackage{nicefrac}
\usepackage{microtype}
\usepackage{graphicx}
\usepackage{amsmath}
\usepackage{amssymb}
\usepackage[table]{xcolor}
\usepackage{multirow}
\usepackage{makecell}
\usepackage{caption}
\usepackage{subcaption}
\usepackage{ragged2e}
\usepackage{algorithm}
\usepackage{algpseudocode}
\usepackage{enumitem}
\usepackage{float}
\usepackage{placeins}
\usepackage{wrapfig}
\usepackage{marvosym}

\definecolor{myblue}{rgb}{0.92, 0.95, 1.0}
\definecolor{ddbfblue}{rgb}{0.92,0.95,1.0}
\newcommand{\dd}[1]{#1}
\newcommand{\ddbf}[1]{\cellcolor{ddbfblue}\textbf{#1}}

\title{GeoProp: Grounding Robot State in Vision for Generalist Manipulation}

\author{
  Guoyang Zhao$^{1,*}$, Quanhao Qian$^{2,3,*}$, Gongjie Zhang$^{4}$, Wenhao Li$^{5}$,\\
  \bfseries Jiuniu Wang$^{2,3}$, Xiaowei Lu$^{2,3}$, Deli Zhao$^{2,3}$, Ran Xu$^{2,3,}$\thinspace$^{\text{\Letter}}$\\
  \mdseries $^{1}$Tongji University \quad $^{2}$DAMO Academy, Alibaba Group \quad $^{3}$HuPan Lab\\
  $^{4}$Alibaba Group \quad $^{5}$Nanyang Technological University\\
  \small $^{*}$Equal contribution. \quad $^{\text{\Letter}}$Corresponding author.
}

\begin{document}
\maketitle

\begin{abstract}
Proprioception is fundamental to robotic manipulation, yet standard fusion methods often treat it as an isolated vector lacking explicit alignment with visual tokens. Without a direct correspondence between 3D kinematics and 2D feature maps, manipulation policies struggle to ground the robot's state within the scene, frequently underperforming even vision-only baselines. To address this, we introduce GeoProp, a lightweight, plug-and-play adapter that aligns proprioception with vision through explicit geometric grounding and spatial feature sampling. GeoProp projects the robot state onto the image plane to sample localized visual features, constructing a grounded state token. It then injects state-derived spatial priors into the corresponding visual features via FiLM modulation. To capture motion intent, GeoProp further samples features at a short-horizon predicted coordinate derived from recent kinematics, providing look-ahead visual context. Across 67 tasks, GeoProp improves Diffusion Policy by 8.7\% on 63 simulation tasks and $\pi_0$ by 4.0\% on the RoboTwin subset, and yields a 10.6\% average gain across both policy families in the real world, while adding only 2--3\% to the parameter count. These results demonstrate that GeoProp is a simple yet high-impact inductive bias for generalist embodied policies. Project page: \url{https://alibaba-damo-academy.github.io/GeoProp/}.
\end{abstract}



\section{Introduction}

\begin{wrapfigure}{r}{0.45\textwidth}
  \vspace{-10pt}
  \centering
  \includegraphics[width=\linewidth]{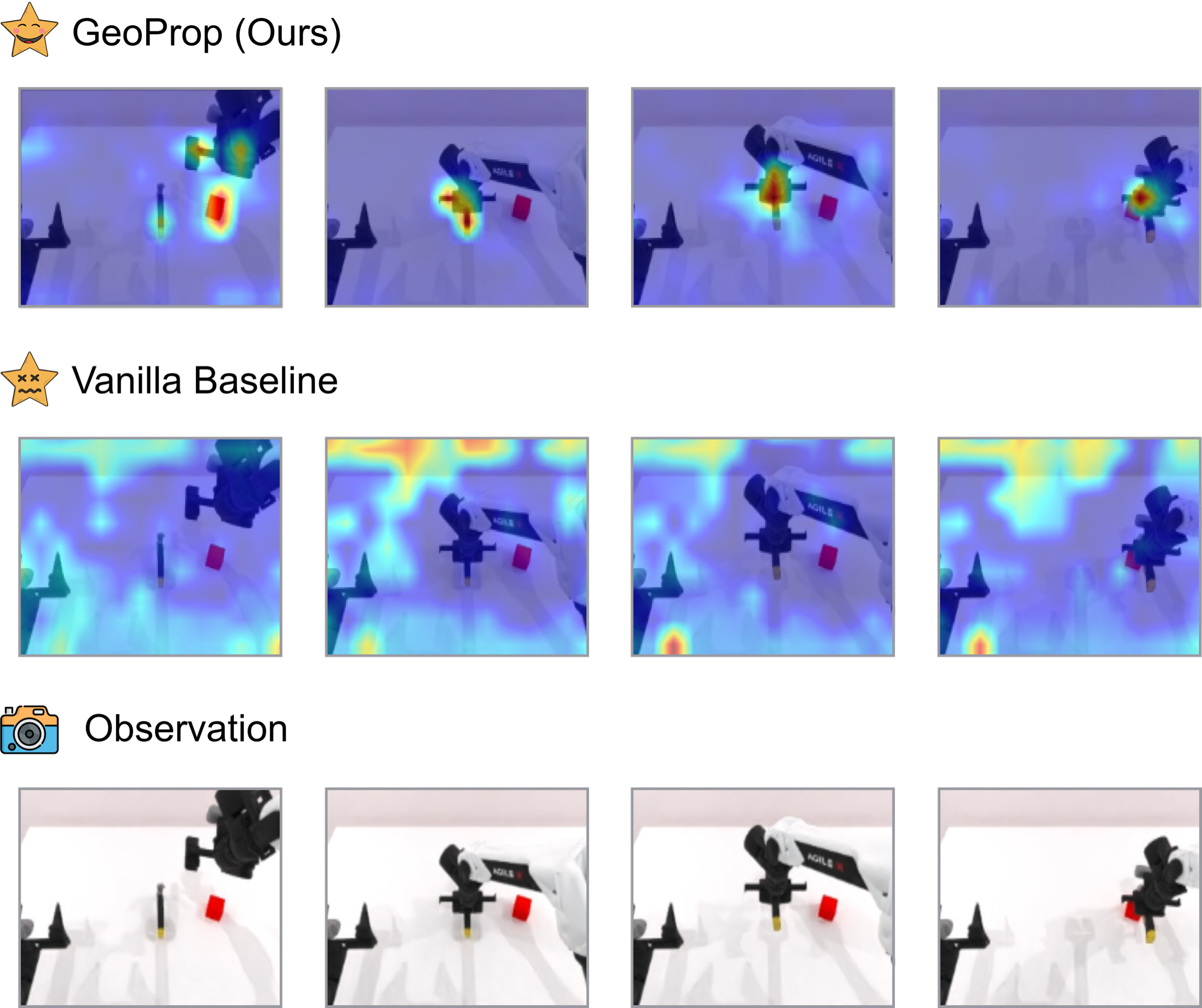}
  \vspace{-8pt}
  \caption{Proprioceptive-to-image attention in $\pi_0$: GeoProp concentrates attention on the gripper and manipulated objects, while vanilla attention is diffuse.}
  \label{fig:overview}
  \vspace{-10pt}
\end{wrapfigure}

Despite the diversification of robot learning architectures---spanning diffusion-based controllers~\citep{chi2023diffusion, ze20243d_3dp, dasari2025ingredients}, transformer-based action predictors~\citep{zhao2023learning, gervet2023act3d}, and large-scale vision--language--action (VLA) systems~\citep{brohan2022rt,zitkovich2023rt, kim2025openvla, black2410pi0}---a persistent representational limitation remains: the structural decoupling between high-dimensional visual observations and low-dimensional proprioceptive feedback. Current mainstream frameworks encode proprioception as a global, ungrounded state embedding and fuse it with vision either through simple concatenation~\citep{jang2022bc, zhao2023learning, kim2025fine_openvla-oft, mees2024octo} or through more expressive cross-attention~\citep{black2410pi0, Nvidia2025GR00TNA}. Both designs typically lack an explicit state-vision correspondence and require the model to learn 3D kinematics-to-2D feature alignment implicitly, overlooking the robot's intrinsic geometric relationship with the visual scene. Our empirical analysis shows that ungrounded proprioception can be counterproductive: without explicit alignment, the state vector may introduce spurious correlations that cause the model to underperform vision-only alternatives.

We propose \textbf{GeoProp}, a lightweight adapter that bridges the gap between 3D kinematics and 2D vision by transforming the robotic state into a localized geometric modulation within the 2D visual feature map.
Instead of treating proprioception as an ungrounded auxiliary input, GeoProp projects the end-effector pose onto the image plane and uses the corresponding localized visual features as the state-aligned visual feature of the robot. This enables the robot's configuration to inherit scene semantics within the same visual latent space, promoting modality-consistent fusion, as visualized by the attention heatmaps in Fig.~\ref{fig:overview} and Appendix~\ref{app:qual_grounding}. In the vanilla $\pi_0$~\citep{black2410pi0} model, state embeddings often fail to produce spatially coherent attention and yield diffuse, background-driven patterns. In contrast, GeoProp tightly couples proprioceptive cues to the relevant visual regions by establishing correspondences analytically.

GeoProp anchors the robotic state within the visual manifold across both static configuration and motion dynamics. First, we employ \textbf{Spatially-Aligned Modulation}---inspired by FiLM~\citep{perez2018film}---to inject state-derived geometric cues into tokens near the projected end-effector location. Second, we introduce \textbf{Predictive Kinematic Sampling} to capture motion intent by sampling visual features at a short-horizon predicted coordinate estimated from recent kinematics, providing a look-ahead visual context.

As GeoProp operates at the interface between perception and state representation, it is inherently framework-agnostic. It addresses a shared limitation of prevailing fusion paradigms: Diffusion Policy~\citep{chi2023diffusion} typically performs shallow state fusion via concatenation, whereas VLA models such as $\pi_0$~\citep{black2410pi0} employ deep cross-modality attention; nevertheless, both commonly represent proprioception as a standalone state vector that is not spatially aligned with visual tokens. GeoProp improves both frameworks by supplying a spatially grounded, geometry-aware signal that aligns robot kinematics with visual semantics, without requiring changes to the backbone architecture. We validate GeoProp across 67 tasks, including 63 tasks in simulation~\citep{yu2020meta, james2020rlbench, mu2025robotwin} and 4 real-world tasks on a Mobile ALOHA system~\citep{fu2024mobilealohalearningbimanual}. Our results show average absolute improvements of \textbf{8.7\%} on Diffusion Policy across 63 simulation tasks and \textbf{4.0\%} on $\pi_0$ on the RoboTwin subset, together with a \textbf{10.6\%} real-world gain averaged over both policy families, with negligible parameter overhead of \textbf{2--3\%}.

\noindent\textbf{Our contributions:} 
(1) We empirically show that treating proprioception as an ungrounded global vector can lead to modality misalignment and degraded manipulation performance compared to vision-only baselines, motivating the need for an explicit, geometry-based state--vision correspondence.
(2) We propose \textbf{GeoProp}, a plug-and-play adapter that transforms 3D proprioception into image-grounded visual tokens, bridging the representational gap between robot kinematics and scene semantics through geometric projection, localized feature modulation, and predictive sampling of motion intent.
(3) We demonstrate that \textbf{GeoProp} consistently enhances performance across distinct policy architectures in both simulation and the real world, validating explicit geometric grounding as an efficient and powerful inductive bias for embodied AI.


\section{Related Work}

\subsection{Generalist Robot Policies and Foundation Models}
Robotic manipulation has shifted from state-based reinforcement learning~\citep{lillicrap2015continuous, schulman2017proximal}, which often assumes compact low-dimensional states, to policies learned from high-dimensional visual observations~\citep{chi2023diffusion, zhao2023learning, majumdar2023we-vc1, shang2024theiadistillingdiversevision}. More recently, Vision-Language-Action (VLA) models~\citep{zitkovich2023rt, kim2025openvla, black2410pi0} leverage large-scale pre-training for semantic generalization and open-vocabulary task following. However, precise manipulation still requires aligning what the robot sees with where the robot physically is. Most frameworks fuse proprioception through concatenation, MLP state encoders, or transformer state/action tokens, treating robot state as a collapsed global context vector. This leaves the correspondence between visual semantics and kinematic state implicit, forcing policies to learn state--vision alignment from data alone.

\subsection{Proprioception--Vision Alignment}
Recent work has examined the limitations of ungrounded proprioceptive fusion. Some studies~\citep{lu2025would, zhao2025you} suggest that isolated state vectors can induce embodiment- or trajectory-specific overfitting; relative-action formulations~\citep{zhao2025you} improve spatial generalization but reduce access to absolute kinematic state. Other methods align proprioception through learned objectives or alternative state parameterizations, including contrastive losses~\citep{kim2025contrastive}, discrete codebooks~\citep{fan2025xr}, and verbalized state tokens~\citep{suzuki2025proprioception}. Most closely related, robot-centric positional encoding (RC-PE)~\citep{li2025towards} projects the end-effector to the image plane and injects a dense relative-coordinate embedding into every visual token. RC-PE shares GeoProp's premise that projecting the robot state into image space exposes a useful geometric cue, but spreads this signal across the full feature grid rather than co-locating proprioception with the visual evidence. A complementary line targets the \emph{input} or \emph{planning} level: LLARVA~\citep{niu2024llarva}, HAMSTER~\citep{li2025hamster}, and PEEK~\citep{zhang2025peek} predict visual traces, paths, or affordance masks from a VLM as intermediate task plans, while AimBot~\citep{dai2025aimbot} overlays analytically-projected robot reticles on the raw RGB input. GeoProp instead operates inside the policy's feature stack, FiLM-modulating intermediate visual \emph{features} at projected end-effector locations without changing the input image or requiring a VLM planner, and is thus complementary to input/planning-level methods.

\subsection{Geometric Grounding in Robotic Perception}
Explicit geometric priors improve sample efficiency, spatial reasoning, and robustness in manipulation~\citep{zheng2025learning, miao2025towards, zhang2026on}. Many methods construct 3D scene representations, including voxel, point-cloud, or 3D-aware policy backbones~\citep{shridhar2023perceiver, goyal2024rvt, ze20243d_3dp, zhang2025flowpolicy}, giving vision and robot state a shared coordinate substrate but often requiring depth, multi-view observations, 3D reconstruction, or dedicated 3D architectures. Recent VLA-oriented methods use 3D Gaussian Splatting or feature distillation for scene-centric spatial reasoning~\citep{li2025spatial, qian2025geopredict, qian2025gp3}; for instance, GeoPredict~\citep{qian2025geopredict} uses predictive 3D Gaussian geometry and robot keypoint trajectories as auxiliary supervision. GeoProp instead takes a lightweight, robot-state-centric route: it anchors end-effector kinematics directly onto 2D visual feature tokens via projective grounding, preserving explicit state--vision correspondence without full 3D reconstruction.

\section{Methodology}

\begin{figure}[t]
    \centering
    \includegraphics[width=0.98\textwidth]{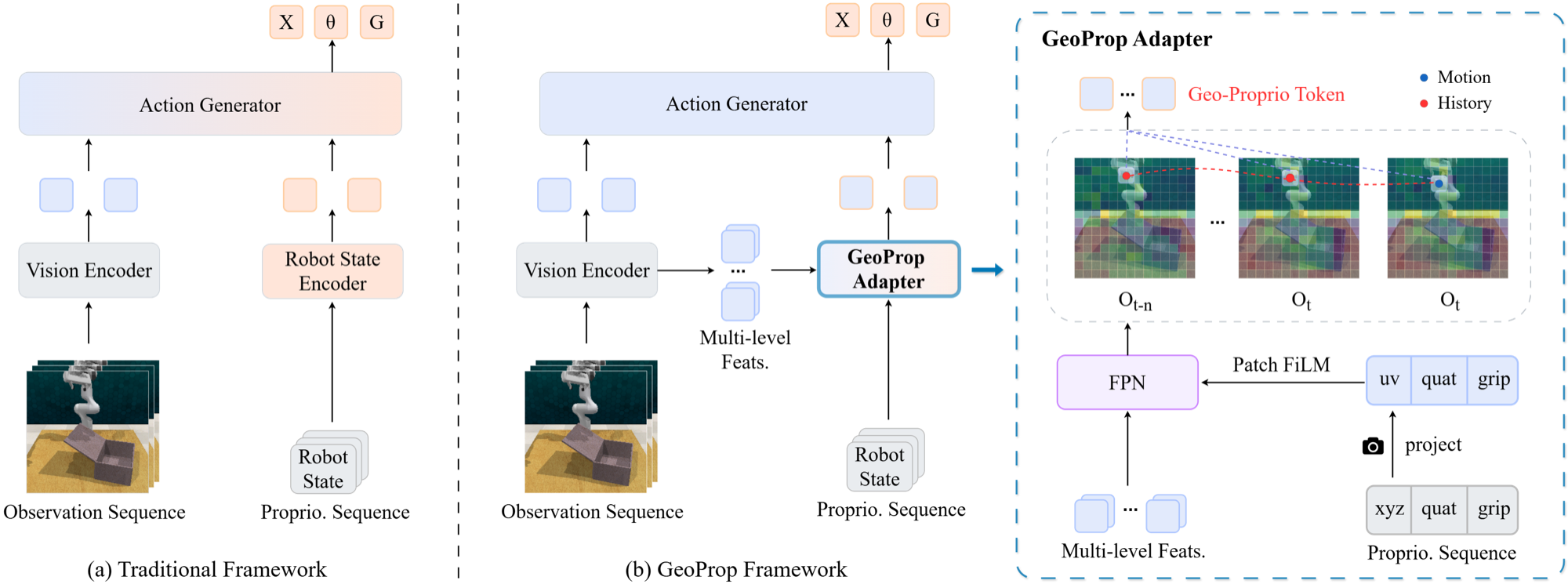}
    \caption{GeoProp projects end-effector and look-ahead waypoints to image features, producing spatially aligned state tokens for downstream policies.}
    \label{fig:overview_arch}
\end{figure}

As illustrated in Fig.~\ref{fig:overview_arch}, GeoProp aligns 3D proprioception with 2D visual tokens through three components: \textbf{i)} geometric projection and feature sampling, \textbf{ii)} spatially aligned feature modulation, and \textbf{iii)} predictive kinematic sampling.
The method assumes access to camera intrinsics $\mathbf{K}$ and extrinsics $(\mathbf{R}, \mathbf{t})$, intermediate spatial feature maps from the vision backbone, and the end-effector 3D position.
The visual token sampled at the projected end-effector location is referred to as the \emph{grounded state token}, as it carries local scene evidence at the robot's current interaction point.
GeoProp uses FPN~\citep{lin2017feature} for multi-scale aggregation and applies FiLM modulation~\citep{perez2018film} only at the aligned feature cell.
The resulting grounded and predictive tokens are concatenated with global visual tokens, and language tokens when available, and passed to the downstream policy without modifying the backbone.

\subsection{Geometric Projection and Feature Grounding}
\label{sec:geom_proj}

We denote the proprioceptive state at time $t$ as
$\mathbf{p}_{t}=[x_t,y_t,z_t,q_{w,t},q_{x,t},q_{y,t},q_{z,t},g_t]^{\top}\in\mathbb{R}^{8}$,
including the end-effector position, orientation, and gripper state.
Standard fusion methods inject $\mathbf{p}_t$ as an ungrounded global vector, requiring the policy to infer the correspondence between 3D kinematics and 2D visual tokens.
GeoProp instead makes this correspondence explicit by projecting the end-effector into the image plane and sampling the visual feature at the projected location.

Let $\mathbf{r}_t=[x_t,y_t,z_t]^\top$ be the end-effector position in the robot/world frame.
Given camera intrinsics $\mathbf{K}$ and extrinsics $(\mathbf{R},\mathbf{t})$, we obtain the image-plane coordinate
\begin{equation}
\label{eq:projection}
\mathbf{q}_t
=
\Pi_{\mathbf{K}}\!\left(\mathbf{R}\mathbf{r}_t+\mathbf{t}\right),
\qquad
\mathbf{q}_t=(u_t,v_t)^\top ,
\end{equation}
where $\Pi_{\mathbf{K}}(\cdot)$ denotes perspective projection with intrinsics $\mathbf{K}$.
The image coordinate $\mathbf{q}_t$ is then mapped to a continuous coordinate
$\bar{\mathbf{q}}_t=\rho(\mathbf{q}_t)$ on the feature grid, where $\rho(\cdot)$ maps image-plane coordinates to the corresponding feature-grid coordinates induced by the visual backbone.

Given a spatial feature map $\mathbf{F}$, the feature at a continuous coordinate is obtained via bilinear sampling:
\begin{equation}
\label{eq:sampling}
\mathcal{S}(\mathbf{F},\bar{\mathbf{q}})
=
\sum_{\mathbf{c}\in\mathcal{N}(\bar{\mathbf{q}})}
w_{\mathbf{c}}(\bar{\mathbf{q}})\,\mathbf{F}[\mathbf{c}],
\end{equation}
where $\mathcal{N}(\bar{\mathbf{q}})$ denotes the four neighboring grid cells and $w_{\mathbf{c}}$ are bilinear interpolation weights.
After the modulation and aggregation stage in Sec.~\ref{sec:film_mod}, GeoProp forms the grounded state token as
\begin{equation}
\label{eq:tau_base}
\boldsymbol{\tau}_{t}
=
\mathcal{S}(\mathbf{F}_{\mathrm{mod}}, \bar{\mathbf{q}}_t).
\end{equation}

\subsection{Spatially-Aligned Feature Modulation}
\label{sec:film_mod}

The grounded token $\boldsymbol{\tau}_t$ provides an image-aligned state representation, but it is extracted only after visual encoding.
To inject the same geometric prior into the visual hierarchy, GeoProp applies state-conditioned FiLM modulation locally at the projected end-effector location before FPN aggregation.

For each intermediate feature map $\mathbf{F}^{\ell}_{\mathrm{enc}}$,
let $\mathbf{c}^{\ell}_t$ denote the aligned grid cell induced by $\bar{\mathbf{q}}_t$ at level $\ell$.
Given the proprioceptive state $\mathbf{p}_t$, GeoProp predicts channel-wise FiLM parameters
$\boldsymbol{\gamma}^{\ell}_t$ and $\boldsymbol{\beta}^{\ell}_t$ for each feature level.
The modulated feature map is defined as
\begin{equation}
\label{eq:film_point}
\mathbf{F}^{\ell}_{\mathrm{mod}}[\mathbf{c}]
=
\begin{cases}
(1+\boldsymbol{\gamma}^{\ell}_t)\odot \mathbf{F}^{\ell}_{\mathrm{enc}}[\mathbf{c}]
+\boldsymbol{\beta}^{\ell}_t,
& \mathbf{c}=\mathbf{c}^{\ell}_t,\\
\mathbf{F}^{\ell}_{\mathrm{enc}}[\mathbf{c}],
& \mathrm{otherwise}.
\end{cases}
\end{equation}
This operation conditions the visual representation only at the feature cell geometrically aligned with the end-effector. Thus, proprioceptive conditioning remains spatially localized prior to multi-scale aggregation.

The modulated multi-scale features are then aggregated with an FPN:
\begin{equation}
\label{eq:fmod}
\mathbf{F}_{\mathrm{mod}}
=
\mathrm{FPN}(\{\mathbf{F}^{\ell}_{\mathrm{mod}}\}_{\ell=1}^{L}).
\end{equation}
The grounded state token is obtained by sampling $\mathbf{F}_{\mathrm{mod}}$ at the continuous feature-grid coordinate defined in Sec.~\ref{sec:geom_proj}.

\subsection{Predictive Kinematic Sampling}
\label{sec:pred_sample}

Instantaneous proprioception localizes the current end-effector state,
but provides limited information about imminent motion trends.
GeoProp therefore augments grounded state representations with a predictive spatial token constructed from a short-horizon look-ahead waypoint.

Given a temporal window of recent 3D positions
$\{\mathbf{r}_{t-k},\dots,\mathbf{r}_t\}$,
we use polynomial extrapolation to estimate a future waypoint
$\hat{\mathbf{r}}_{t+1}$.
This waypoint is projected to the image plane using Eq.~\eqref{eq:projection} and mapped to the feature grid, yielding a continuous look-ahead coordinate
$\bar{\mathbf{q}}^{\mathrm{pre}}_{t+1}$.

Let $\mathbf{F}_{\mathrm{raw}}$ denote the FPN-aggregated feature map computed from the unmodulated encoder features.
We construct the predictive token by sampling this feature map at the look-ahead coordinate:
\begin{equation}
\label{eq:tau_pre}
\boldsymbol{\tau}_{\mathrm{pre}}
=
\mathcal{S}(\mathbf{F}_{\mathrm{raw}}, \bar{\mathbf{q}}^{\mathrm{pre}}_{t+1}).
\end{equation}
Sampling from unmodulated features decouples the look-ahead token from state-conditioned modulation, making it a spatial preview of the anticipated motion rather than an additional modulation site.

Let $\mathbf{h}_{\mathrm{global}}$ denote the set of global visual tokens produced by the vision backbone.
The policy input is formed by concatenating the grounded state token, the predictive token, and the global visual tokens:
\begin{equation}
\label{eq:policy_input}
\mathbf{z}_t
=
[\boldsymbol{\tau}_t;\boldsymbol{\tau}_{\mathrm{pre}};\mathbf{h}_{\mathrm{global}}].
\end{equation}
This representation jointly encodes grounded current state, anticipated motion context, and global scene semantics within a unified geometry-aware feature space. For bimanual platforms (e.g., the ALOHA setups in our RoboTwin and Mobile ALOHA experiments), GeoProp is applied independently to each arm, producing one grounded and one predictive token per end-effector that are concatenated with the global visual tokens.


\section{Simulation Experiments}
\label{sec:sim_exp}

We evaluate GeoProp on 63 simulated tasks across three benchmarks spanning single-arm and bimanual manipulation, relative and absolute pose control, and both task-specific and multi-task training.

\subsection{Experimental Setup}
\textbf{Benchmarks.}
We evaluate on three established manipulation benchmarks.
MetaWorld~\citep{yu2020meta} contains 50 tabletop manipulation tasks with a Sawyer robot in MuJoCo; we train one policy per task using 25 scripted demonstrations and a corner-view camera, with actions parameterized as end-effector translation deltas $(\Delta x,\Delta y,\Delta z)$. We follow~\citet{qian2025gp3} to group the 50 tasks into the Easy / Medium / Hard / Very~Hard difficulty splits used in Table~\ref{tab:dp_results}.
RLBench~\citep{james2020rlbench} provides visually rich manipulation tasks in CoppeliaSim with a Franka Panda; following prior work~\citep{jia2025lift3d,qian2025gp3}, we evaluate on six tasks using 100 OMPL-generated demonstrations per task, front-view images, and absolute 7-DoF end-effector pose actions.
RoboTwin~\citep{mu2025robotwin} evaluates bimanual manipulation with a simulated ALOHA setup; following~\citep{fang2025robotic}, we use seven tasks with 50 demonstrations per task, head-camera observations, and absolute 14-DoF actions for the two arms.
MetaWorld uses per-task training, while RLBench and RoboTwin use multi-task training. 

\textbf{Baselines.}
We compare GeoProp with two policy families that represent different mechanisms for proprioception--vision integration.
Diffusion Policy (DP)~\citep{chi2023diffusion} uses shallow fusion, where proprioception is encoded by an MLP and concatenated with visual tokens for policy conditioning.
We instantiate DP with both ResNet-18~\citep{he2016deep} and ViT-B/16~\citep{dosovitskiy2020vit} vision backbones to test backbone generality.
We also evaluate $\pi_0$~\citep{black2410pi0}, a 3B-parameter vision-language-action model in which state, image, and language tokens interact through deep transformer attention.
For each policy family, we compare three variants: \textbf{i)} No-Proprio, which removes proprioceptive input; \textbf{ii)} Vanilla, which uses the original proprioceptive conditioning mechanism; and \textbf{iii)} GeoProp, which replaces the original proprioceptive conditioning mechanism with our geometry-grounded adapter.
All variants use the same data, backbone, action space, and training schedule within each policy family. This setup isolates the effect of proprioceptive conditioning while controlling for backbone capacity, action parameterization, and optimization settings.

\textbf{Implementation details.}
All policies process RGB observations resized to $224{\times}224$. For success-rate reporting, MetaWorld results are averaged over $2$ seeds with $25$ evaluation rollouts each ($50$ rollouts per task), while RLBench and RoboTwin use $25$ evaluation rollouts per task.
For DP, ResNet-18 is trained from scratch, while ViT-B/16 is initialized from ImageNet-1K pretrained weights using \texttt{timm}.
DP is trained for 100 epochs with Adam $(\beta_1,\beta_2)=(0.95,0.999)$, a cosine learning-rate schedule, 10\% linear warm-up, and initial learning rate $1{\times}10^{-4}$.
For $\pi_0$, we fine-tune for 30k steps with AdamW $(\beta_1,\beta_2)=(0.9,0.95)$, a cosine schedule, 3\% warm-up, and peak learning rate $5{\times}10^{-5}$.
GeoProp uses two-layer MLPs to generate FiLM parameters, and predictive kinematic sampling extrapolates one future waypoint from a 4-frame history window.
Full architectural and hyperparameter details are provided in Appendix~\ref{sec:implementation_details}.

\begin{table*}[t]
\centering
\caption{Diffusion Policy success rates (\%) across simulation benchmarks. Results are reported with ResNet-18 and ViT-Base backbones. Per-task numbers are in Appendix~\ref{app:per_task_sim}. Parentheses denote absolute gains over Vanilla, and $\Delta$ Params reports the parameter overhead of GeoProp.}
\label{tab:dp_results}
\small
\setlength{\tabcolsep}{3.4pt}
\begin{tabular}{@{}ll cccc cc c c@{}}
\toprule
\multirow{2}{*}{\textbf{Backbone}} &
\multirow{2}{*}{\textbf{Method}} &
\multicolumn{4}{c}{\textbf{MetaWorld (50)}} &
\textbf{RLBench} &
\textbf{RoboTwin} &
\textbf{Overall} &
\multirow{2}{*}{\textbf{\makecell{$\Delta$ Params}}} \\
\cmidrule(lr){3-6} \cmidrule(lr){7-7} \cmidrule(lr){8-8} \cmidrule(lr){9-9}
& & Easy & Medium & Hard & V. Hard & Mean & Mean & Avg. & \\
\midrule

\multirow{3}{*}{ResNet-18}
& Vanilla    & 81.8 & 49.1 & 46.0 & 66.4 & 66.0 & 53.7 & 66.8 & -- \\
& No-Proprio & 80.9 & 51.1 & 51.0 & 74.4 & 68.0 & 53.7 & 68.1 & -- \\
\rowcolor{myblue}
\cellcolor{white} & \textbf{GeoProp} & \textbf{82.9} & \textbf{67.5} & \textbf{59.0} & \textbf{82.4} & \textbf{79.3} & \textbf{62.9} & \textbf{75.3} (\textbf{+8.5}) & \textbf{+3.0\%} \\

\midrule

\multirow{3}{*}{ViT-Base}
& Vanilla    & 78.3 & 46.7 & 34.0 & 63.6 & 65.3 & 41.1 & 62.0 & -- \\
& No-Proprio & 78.5 & 50.5 & 30.7 & 52.8 & 70.7 & 42.9 & 62.3 & -- \\
\rowcolor{myblue}
\cellcolor{white} & \textbf{GeoProp} & \textbf{81.5} & \textbf{62.5} & \textbf{52.7} & \textbf{69.2} & \textbf{79.3} & \textbf{52.0} & \textbf{71.0} (\textbf{+9.0}) & \textbf{+2.3\%} \\

\bottomrule
\end{tabular}
\end{table*}

\subsection{Quantitative Results}

Table~\ref{tab:dp_results} reports the main results on Diffusion Policy (DP).
Across 63 tasks and two vision backbones, GeoProp achieves the best overall performance.
With ResNet-18, GeoProp reaches 75.3\% average success, improving over Vanilla by 8.5 points and over No-Proprio by 7.2 points.
With ViT-Base, GeoProp achieves 71.0\%, yielding 9.0 and 8.7 point gains over Vanilla and No-Proprio, respectively.
The gains are especially pronounced on the harder MetaWorld subsets, suggesting that spatially grounded proprioception is more beneficial in manipulation settings that require precise visual-state alignment.
These improvements come with limited overhead, adding only 2.3--3.0\% parameters to DP.

\begin{table}[t]
\centering
\caption{Success rates (\%) on $\pi_0$ across RoboTwin tasks.}
\label{tab:pi0_comparison}
\small
\setlength{\tabcolsep}{4pt}
\begin{tabular}{@{}lccccccc c@{}}
\toprule
\textbf{Method} & \makecell{\textbf{Beat Block}\\\textbf{Hammer}} & \makecell{\textbf{Click Alarm}\\\textbf{Clock}} & \makecell{\textbf{Move Can}\\\textbf{to Pot}} & \makecell{\textbf{Move Playing}\\\textbf{Card}} & \makecell{\textbf{Place}\\\textbf{Shoe}} & \makecell{\textbf{Scan}\\\textbf{Object}} & \makecell{\textbf{Stack Two}\\\textbf{Blocks}} & \textbf{Avg.} \\
\midrule
No-Proprio & 80 & 88 & 80 & \textbf{76} & 40 & 24 & 44 & 61.7 \\
$\pi_0$ (Vanilla) & 76 & \textbf{100} & 80 & 72 & 48 & 24 & 48 & 64.0 \\
\rowcolor{myblue}
\textbf{GeoProp} & \textbf{84} & 88 & \textbf{92} & 72 & \textbf{52} & \textbf{28} & \textbf{60} & \textbf{68.0} \\
\bottomrule
\end{tabular}
\end{table}

We further evaluate GeoProp on the VLA model $\pi_0$ in Table~\ref{tab:pi0_comparison}.
Since $\pi_0$ uses deep cross-modality attention between state, image, and language tokens, it provides a strong baseline for implicit proprioception--vision alignment.
GeoProp improves the average success rate from 64.0\% to 68.0\% and outperforms the vanilla $\pi_0$ baseline on 5 out of 7 tasks.
This suggests that explicit geometric grounding remains complementary even when the backbone has substantial capacity for cross-modal interaction.
The parameter overhead on $\pi_0$ is also small (+1.98\%).
The gain on $\pi_0$ further holds across data scales (+2.3 / +4.0 / +2.8 pp at 25 / 50 / 100 demos per task) and on an expanded 15-task RoboTwin suite (+4.5pp); full results are reported in Appendix~\ref{app:additional_quant}.

Across DP backbones in simulation, Vanilla proprioception fusion is often close to or below No-Proprio.
This indicates that ungrounded state vectors are not always used effectively by the policy.
By anchoring state information to the corresponding visual feature location, GeoProp makes proprioception more consistently useful at the aggregate level.

\subsection{Ablations and Analysis}
\label{sec:ablations}

We conduct ablations and diagnostic analyses to identify which design choices drive GeoProp's gains and when the geometric grounding assumption is most beneficial.
Specifically, we study \textbf{i)} component contributions, \textbf{ii)} whether simpler spatial-coordinate priors can explain the improvement, and \textbf{iii)} task-type patterns and robustness to camera--robot calibration errors.

\begin{wraptable}{r}{0.48\textwidth}
\vspace{0.5em}
\centering
\caption{Component ablation on 15 MetaWorld tasks. Spa.: geometric grounding; Mot.: predictive sampling; FiLM: localized modulation.}
\label{tab:ablation_components}
\footnotesize
\resizebox{\linewidth}{!}{%
\setlength{\tabcolsep}{3pt}%
\begin{tabular}{@{}lccccc@{}}
\toprule
\textbf{Component} & \textbf{Vanilla} & \textbf{Base} & \textbf{+\,Mot.} & \textbf{+\,FiLM} & \cellcolor{myblue}\textbf{Full} \\
\midrule
Spa.\ grounding   & -- & \checkmark & \checkmark & \checkmark & \cellcolor{myblue}\checkmark \\
Mot.\ sampling    & -- & --         & \checkmark & --         & \cellcolor{myblue}\checkmark \\
FiLM modulation   & -- & --         & --         & \checkmark & \cellcolor{myblue}\checkmark \\
\midrule
\textbf{Mean (\%)} & 61.2 & 64.8 & 66.4 & 66.0 & \cellcolor{myblue}\textbf{68.5} \\
\bottomrule
\end{tabular}}%
\vspace{-1em}
\end{wraptable}
\noindent\textbf{Component Effectiveness.}
We ablate three components of GeoProp: geometric spatial grounding (Spa.), predictive kinematic sampling (Mot.), and spatially-aligned FiLM modulation.
Following~\citep{qian2025gp3}, we evaluate all variants on 15 diverse MetaWorld tasks.
As shown in Table~\ref{tab:ablation_components}, replacing vector-based proprioception fusion with geometric grounding improves mean success from 61.2\% to 64.8\%.
Adding predictive sampling and localized FiLM further improves performance to 66.4\% and 66.0\%, respectively, while the full model reaches 68.5\%.
These results indicate that geometric grounding is the primary source of improvement, with motion look-ahead and localized modulation providing complementary gains. Notably, motion sampling and FiLM modulation are super-additive (\textbf{+3.7pp} jointly vs.\ +1.6 / +1.2 individually), suggesting they address complementary error modes.
Split-level results are provided in Appendix~\ref{app:additional_quant}.

\textbf{Alternative Spatial-Grounding Baselines.}
To test whether GeoProp's gains come merely from providing the projected end-effector location, we compare against Heatmap injection and RC-PE~\citep{li2025towards}.
On the same 15 MetaWorld tasks, Heatmap and RC-PE reach 61.7\% and 60.8\% mean success, respectively, compared with 59.5\% for No-Proprio and 64.8\% for GeoProp-Base.
Thus, simple coordinate priors help, but do not match co-located visual feature sampling.
Full split-level results are provided in Appendix~\ref{app:additional_quant}.

\textbf{Task-Type Patterns.}
GeoProp provides the largest gains on precision-oriented tasks involving small objects: on the four representative MetaWorld tasks \textit{Basketball}, \textit{Hand Insert}, \textit{Pick Place}, and \textit{Sweep Into}, the mean improvement reaches \textbf{+22.3pp} averaged over both backbones (Appendix Table~\ref{tab:app_smallobj_precision}).
GeoProp underperforms when the manipulated object occludes the projected end-effector, e.g., on \emph{Box Close} (RN: 48$\rightarrow$40; ViT: 54$\rightarrow$54), where the lid covers the gripper during closing.

\textbf{Parameter-Matched Control.} A parameter-matched control that enlarges the proprioceptive encoder to GeoProp's parameter count recovers only \textbf{+0.5pp} for DP and \textbf{+0.1pp} for $\pi_0$, indicating the gains are driven primarily by geometric alignment rather than added capacity.

\textbf{Calibration Robustness.}
GeoProp relies on camera--robot calibration for projection, so we stress-test it under synthetic extrinsic perturbations at test time.
As detailed in Appendix~\ref{app:calibration_robustness}, GeoProp degrades gradually under moderate calibration drift and remains above Vanilla and No-Proprio for translation errors up to 2.5\,cm and rotation errors up to $1.5^\circ$.


\section{Real-World Experiments}
\label{sec:real_world}

\subsection{Experimental Setup}

We evaluate GeoProp on the Mobile ALOHA bimanual platform~\citep{fu2024mobilealohalearningbimanual}, equipped with two 6-DoF arms and a front-facing Intel RealSense camera.
The setup reflects a common single-view deployment scenario where proprioception must be aligned with egocentric visual observations under sensor noise, mild calibration drift, and cluttered household scenes.

We consider four household manipulation tasks that cover distinct interaction patterns:
\textbf{i)} Paper Toss, where the robot grasps a paper ball and throws it into a bin;
\textbf{ii)} Coffee Retrieval, which requires reaching, grasping, and placing a cup;
\textbf{iii)} Desk Clearing, where the robot pushes multiple objects into a target region; and
\textbf{iv)} Table Cleaning, which requires wiping a designated area with a cloth.
Each task uses 50 successful teleoperated demonstrations, and success rates are averaged over 20 evaluation trials per task.
We compare No-Proprio, Vanilla, and GeoProp using Diffusion Policy with a ResNet-18 backbone, while keeping the demonstrations, camera setup, action representation, and training schedule fixed.
We additionally evaluate $\pi_0$ under the same real-world protocol to test whether the benefit of geometric grounding transfers beyond diffusion policies.
Qualitative execution sequences are provided in Appendix~\ref{sec:real_world_qualitative}.

\subsection{Performance Analysis}

\begin{wraptable}{r}{0.52\textwidth}
\vspace{0.5em}
\centering
\caption{Real-world success rates (\%) on Mobile ALOHA. All methods use the same demonstrations, camera setup, and evaluation protocol.}
\label{tab:real_world_results}
\footnotesize
\setlength{\tabcolsep}{2.8pt}
\begin{tabular}{@{}llccccc@{}}
\toprule
\textbf{Policy} & \textbf{Method} & \textbf{Paper} & \textbf{Coffee} & \textbf{Desk} & \textbf{Table} & \textbf{Avg.} \\
\midrule
\multirow{3}{*}{DP}
& No-Proprio & 25 & 40 & 35 & 15 & 28.8 \\
& Vanilla & 35 & 35 & 55 & 30 & 38.8 \\
\rowcolor{myblue}
& \textbf{GeoProp} & \textbf{50} & \textbf{55} & \textbf{60} & \textbf{35} & \textbf{50.0} \\
\midrule
\multirow{3}{*}{$\pi_0$}
& No-Proprio & 45 & 40 & 60 & 25 & 42.5 \\
& Vanilla & 40 & 45 & 55 & 35 & 43.8 \\
\rowcolor{myblue}
& \textbf{GeoProp} & \textbf{55} & \textbf{55} & \textbf{65} & \textbf{40} & \textbf{53.8} \\
\bottomrule
\end{tabular}
\vspace{-1em}
\end{wraptable}
Table~\ref{tab:real_world_results} reports real-world results for both policy families.
GeoProp achieves the highest average success rate, improving from 38.8\% with Vanilla to 50.0\% (\textbf{+11.2pp}) and from 28.8\% with No-Proprio to 50.0\% (\textbf{+21.2pp}).
The gains are largest on tasks that require accurate local contact and gripper--object alignment.
For example, GeoProp improves Coffee Retrieval from 35\% to 55\%, corresponding to four additional successful trials out of 20, and improves Paper Toss from 35\% to 50\% under varied object placements.

The real-world results also reveal the brittleness of ungrounded state fusion.
On Coffee Retrieval, Vanilla underperforms the vision-only baseline (35\% vs.\ 40\%), suggesting that concatenated proprioceptive vectors can interfere with visual policy learning when they are not spatially aligned.
By projecting the end-effector state into the image and sampling co-located visual evidence, GeoProp provides a geometric inductive bias that stabilizes hand-eye coordination under real-world noise factors such as sensor jitter, clutter, and mild calibration drift.

GeoProp shows a similar trend with $\pi_0$ under the same demonstration and evaluation protocol.
It improves average success from 43.8\% to 53.8\% over Vanilla $\pi_0$ (\textbf{+10.0pp}), with per-task gains of \textbf{+15/+10/+10/+5pp} on Paper Toss, Coffee Retrieval, Desk Clearing, and Table Cleaning, respectively.
These results suggest that GeoProp's real-world benefit is not tied to a single policy family, but transfers to both diffusion-based policies and VLA models.


\section{Conclusion and Limitations}

We introduce \textbf{GeoProp}, a lightweight adapter that anchors proprioception to co-located visual semantics by analytically projecting 3D end-effector states into 2D visual feature maps. Across 67 tasks, GeoProp improves Diffusion Policy by \textbf{+8.7\%} and $\pi_0$ by \textbf{+4.0\%} in simulation, and yields a \textbf{+10.6\%} real-world gain, with only \textbf{2--3\%} parameter overhead.
\textbf{Limitations.} GeoProp requires known camera intrinsics and extrinsics: it degrades gradually under mild drift (Appendix~\ref{app:calibration_robustness}) but settings without reliable calibration require re-estimating the camera pose. It grounds only the 3D end-effector position, leaving richer kinematic structures (links, joints, fingertips, contact patches) unrepresented, which limits applicability to dexterous or whole-body manipulation. Predictive sampling further assumes locally smooth end-effector motion; abrupt direction changes or contact transitions can produce off-task look-ahead points. Gains shrink or invert when the manipulated object occludes the projected pixel or when the baseline is already near ceiling; all evaluations also use a single fixed primary camera per scene, leaving multi-view fusion and moving-camera setups open. Finally, our evaluation covers three embodiments and two imitation-learning policy families; transfer to humanoids, dexterous hands, or RL, along with calibration-aware alignment and multi-point kinematic grounding, is left to future work.


\clearpage
\acknowledgments{Acknowledgments omitted for anonymous review.}


\bibliography{references}

\appendix

\makeatletter
\@addtoreset{figure}{section}
\@addtoreset{table}{section}
\makeatother
\renewcommand{\thefigure}{\thesection.\arabic{figure}}
\renewcommand{\thetable}{\thesection.\arabic{table}}

\section{Implementation Details}
\label{sec:implementation_details}

\subsection{Training Setup and Hyperparameters}

\textbf{Visual encoders and feature aggregation.}
GeoProp is backbone-agnostic and only requires spatial feature maps that preserve image-grid correspondence.
To make projection-based sampling comparable across backbones, we use an FPN to produce a dense sampling map with a unified spatial resolution.
For ResNet-18, we aggregate features from the four residual stages and use a stride-16 FPN output as the sampling map.
For ViT-based backbones, including ViT-Base and the SigLIP encoder in $\pi_0$, we extract intermediate transformer features, reshape patch tokens to spatial grids, and aggregate them with the same FPN interface.
This produces a $14{\times}14$ dense feature map for $224{\times}224$ inputs.

\textbf{Localized FiLM modulation.}
The FiLM generator maps the proprioceptive state to channel-wise modulation parameters through a two-layer MLP.
Following Sec.~\ref{sec:film_mod}, FiLM is applied before FPN aggregation and only at the feature cell aligned with the projected end-effector location.
We use the residual FiLM form
\[
\mathbf{F}^{\ell}_{\mathrm{mod}}[\mathbf{c}]
=
(1+\boldsymbol{\gamma}^{\ell}_t)\odot \mathbf{F}^{\ell}_{\mathrm{enc}}[\mathbf{c}]
+
\boldsymbol{\beta}^{\ell}_t,
\quad \mathbf{c}=\mathbf{c}^{\ell}_t ,
\]
while all non-aligned cells remain unchanged.
The last layer of the FiLM generator is zero-initialized, so the initial modulation is an identity transform.
The modulated multi-scale features are then aggregated by the FPN to form $\mathbf{F}_{\mathrm{mod}}$, from which the grounded state token is extracted by bilinear sampling.

\textbf{Predictive kinematic extrapolation.}
To estimate short-horizon motion intent, we fit a quadratic polynomial to the recent end-effector positions along each Cartesian axis and extrapolate one future waypoint.
The predicted waypoint is projected to the image plane and mapped to the feature grid, yielding $\bar{\mathbf{q}}^{\mathrm{pre}}_{t+1}$.
The predictive token is sampled from an FPN-aggregated but non-modulated feature map:
\[
\mathbf{F}_{\mathrm{raw}}
=
\mathrm{FPN}(\{\mathbf{F}^{\ell}_{\mathrm{enc}}\}_{\ell=1}^{L}),
\qquad
\boldsymbol{\tau}_{\mathrm{pre}}
=
\mathcal{S}(\mathbf{F}_{\mathrm{raw}}, \bar{\mathbf{q}}^{\mathrm{pre}}_{t+1}).
\]
Thus, both tokens are extracted from FPN outputs with the same spatial resolution: $\boldsymbol{\tau}_t$ from the modulated map $\mathbf{F}_{\mathrm{mod}}$, and $\boldsymbol{\tau}_{\mathrm{pre}}$ from the unmodulated map $\mathbf{F}_{\mathrm{raw}}$.

\textbf{Implementation and training protocols.}
All policies are trained using standard visuomotor imitation learning setups.
Diffusion Policy is optimized with Adam and a cosine schedule, while $\pi_0$ follows its native fine-tuning protocol with AdamW and linear warm-up.
Detailed architectural and optimization settings are summarized in Table~\ref{tab:hyperparams_detailed}.

\begin{table}[htbp]
\centering
\small
\setlength{\tabcolsep}{7pt}
\caption{Detailed training hyperparameters.}
\label{tab:hyperparams_detailed}
\begin{tabular}{lcc}
\toprule
\textbf{Hyperparameter} & \textbf{Diffusion Policy (DP)} & \textbf{$\pi_0$ VLA} \\
\midrule
Input Resolution & $224 \times 224$ & $224 \times 224$ \\
Image Backbone & ResNet-18 / ViT-B & SigLIP-So400M \\
Observation Horizon & 2 & 1 \\
Action Chunk Steps & 32 & 50 \\
\midrule
ViT Selected Layers & $[2, 5, 8, 11]$ & $[5, 12, 19, 26]$ \\
FPN Target Output & $c4$ (ResNet) / $c3$ (ViT) & $c3$ \\
Feature Map Size & $14 \times 14$ & $14 \times 14$ \\
Feature Dim & 512 (ResNet) / 768 (ViT) & 2048 \\
Kinematic History & 4 & 4 \\
Extrapolation Order & Quadratic & Quadratic \\
\midrule
Training Duration & 100 epochs & 30k steps \\
Optimizer & Adam & AdamW \\
Base Learning Rate & $1 \times 10^{-4}$ & $5 \times 10^{-5}$ \\
Batch Size & 64 & 32 \\
Weight Decay & $1 \times 10^{-6}$ & $0.05$ \\
Learning Rate Schedule & Cosine annealing & Cosine with warm-up \\
\bottomrule
\end{tabular}
\end{table}

\subsection{Baseline Configurations and Fairness}
\label{sec:baseline_fairness}

\textbf{Fair comparison protocol.}
For all comparisons, we keep the dataset split, image preprocessing, visual/language backbone, policy architecture, action representation, rollout horizon, optimizer, and training budget fixed.
The only changed factor is the mechanism used to incorporate proprioception.

\textbf{Vanilla proprioception fusion.}
For Diffusion Policy, Vanilla encodes the proprioceptive vector with a two-layer MLP and concatenates the resulting state embedding with the visual feature before policy conditioning.
For $\pi_0$, proprioception is projected into state tokens in the action expert, while image and language tokens are produced by the VLM backbone.
These tokens interact through the standard transformer attention layers, following the original $\pi_0$ proprioception-conditioning design.
Unlike GeoProp, Vanilla does not impose an explicit correspondence between the robot state and image-space visual tokens.

\textbf{No-Proprio.}
No-Proprio removes the state encoder and trains the policy using only image inputs, and language inputs when available.
This baseline measures whether conventional proprioception fusion provides benefit beyond visual observation alone under the same training budget.

\begin{figure}[htbp]
    \centering
    \includegraphics[width=0.62\linewidth]{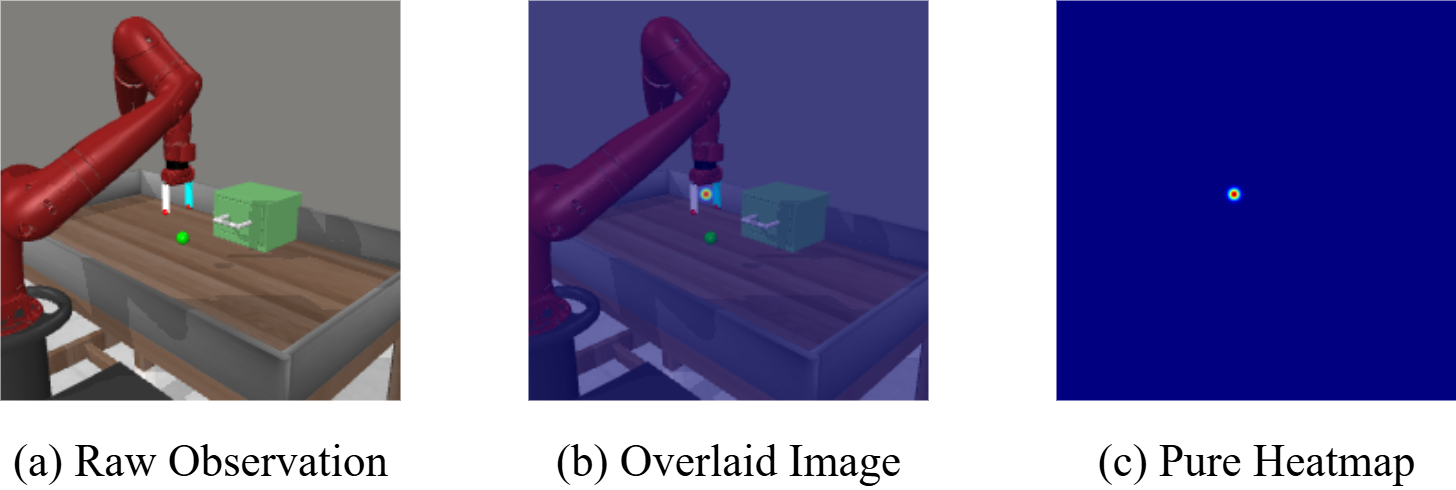}
    \caption{Heatmap conditioning baseline. The projected end-effector coordinate is rendered as a Gaussian heatmap and concatenated with the RGB observation as additional input channels. The overlay is shown only for visualization.}
    \label{fig:heatmap_vis}
\end{figure}

\textbf{Heatmap conditioning.}
Heatmap conditioning provides a spatially explicit but input-level proprioceptive prior.
We project the 3D end-effector position to the image plane and render a Gaussian heatmap centered at the projected pixel.
The heatmap is concatenated with the RGB observation as additional input channels:
\[
I'_t = \mathrm{Concat}(I_t,H_t).
\]
All downstream architecture and training settings are unchanged.
This baseline tests whether exposing the projected state location in image space is sufficient, without co-located feature sampling or localized modulation.

\textbf{RC-PE.}
RC-PE injects a dense robot-centric positional encoding into visual tokens.
For each feature-grid location $(x,y)$, we compute its normalized 2D offset to the projected end-effector coordinate $(\bar{u},\bar{v})$:
\begin{equation}
\Delta x = \frac{x-\bar{u}}{w/2},
\qquad
\Delta y = \frac{y-\bar{v}}{h/2},
\end{equation}
where $w$ and $h$ are the spatial dimensions of the feature grid.
The offset $(\Delta x,\Delta y)$ is embedded with a positional encoding function and added to the corresponding visual token.
This baseline tests whether dense relative-coordinate cues can replace GeoProp's localized feature grounding.

\section{Simulation Experiments Details}
\label{app:sim_details}

\subsection{Additional Quantitative Results}
\label{app:additional_quant}

This section provides additional quantitative results supporting the ablation and scalability analyses in the main paper.
We include split-level component ablations, comparisons with alternative spatial-grounding baselines, and data-scaling results on RoboTwin.

\textbf{Component ablation.}
Table~\ref{tab:app_ablation_components} reports split-level ablations on 15 MetaWorld tasks.
Replacing vector-based proprioception fusion with geometric spatial grounding improves the mean success rate from 61.2\% to 64.8\%.
Adding predictive kinematic sampling and localized FiLM modulation further improves mean success to 66.4\% and 66.0\%, respectively.
The full GeoProp model achieves the best mean success rate of 68.5\%, indicating that geometric grounding is the primary source of improvement, while motion look-ahead and localized modulation provide complementary benefits.

\begin{table}[!htbp]
\centering
\caption{Split-level component ablation on 15 MetaWorld tasks.}
\label{tab:app_ablation_components}
\small
\setlength{\tabcolsep}{3.0pt}
\begin{tabular}{@{}lccc cccc@{}}
\toprule
& \multicolumn{3}{c}{\textbf{Components}} & \multicolumn{4}{c}{\textbf{S.R. (\%)}} \\
\cmidrule(lr){2-4} \cmidrule(l){5-8}
\textbf{Method} & \textbf{Spa.} & \textbf{Mot.} & \textbf{FiLM} & \textbf{Easy} & \textbf{Med.} & \textbf{Hard} & \textbf{Mean} \\
\midrule
Vanilla & -- & -- & --
& $58.3_{\pm3.2}$ & $64.4_{\pm10.7}$ & $62.0_{\pm4.7}$ & 61.2 \\
GeoProp-Base & \checkmark & -- & --
& $57.1_{\pm0.8}$ & $75.6_{\pm5.1}$ & $64.7_{\pm0.9}$ & 64.8 \\
+ Motion & \checkmark & \checkmark & --
& $57.2_{\pm1.6}$ & $\mathbf{80.0}_{\pm4.5}$ & $66.1_{\pm2.4}$ & 66.4 \\
+ FiLM & \checkmark & -- & \checkmark
& $56.9_{\pm1.2}$ & $76.8_{\pm2.3}$ & $69.3_{\pm3.8}$ & 66.0 \\
\rowcolor{myblue}
\textbf{GeoProp-Full} & \checkmark & \checkmark & \checkmark
& $\mathbf{60.3}_{\pm1.2}$ & $78.4_{\pm0.0}$ & $\mathbf{71.3}_{\pm1.0}$ & \textbf{68.5} \\
\bottomrule
\end{tabular}
\end{table}

\textbf{Alternative spatial-grounding baselines.}
To test whether GeoProp's gains come merely from exposing the projected end-effector location, we compare against Heatmap conditioning and RC-PE.
Heatmap renders the projected end-effector coordinate as an image-space Gaussian prior, while RC-PE injects relative coordinate encodings into visual tokens.
As shown in Table~\ref{tab:app_ablation_alternatives}, both baselines improve over No-Proprio, confirming that robot-centric spatial cues are useful.
However, GeoProp-Base reaches 64.8\% mean success, outperforming Heatmap by 3.1 points and RC-PE by 4.0 points.
This suggests that GeoProp's gains come not only from providing location cues, but from grounding proprioception through co-located visual feature sampling.

\textbf{Data scaling.}
Beyond the 50-demo setting reported in the main text, we evaluate whether GeoProp's benefit holds across data scales on the original 7-task RoboTwin setup with $\pi_0$ (Table~\ref{tab:app_data_scaling}).
GeoProp improves over Vanilla by \textbf{+2.3 / +4.0 / +2.8} points at 25 / 50 / 100 demonstrations per task.
The absolute gap peaks in the mid-data regime and narrows as data grows, consistent with the role of geometric grounding as an inductive bias that is most useful when the policy cannot easily infer state--vision alignment from data alone.

\begin{table}[!htbp]
\centering
\caption{Alternative spatial-grounding baselines on 15 MetaWorld tasks.}
\label{tab:app_ablation_alternatives}
\small
\setlength{\tabcolsep}{5.0pt}
\begin{tabular}{@{}lcccc@{}}
\toprule
\textbf{Method} & \textbf{Easy} & \textbf{Med.} & \textbf{Hard} & \textbf{Mean} \\
\midrule
No-Proprio & $54.6_{\pm1.2}$ & $66.0_{\pm1.7}$ & $58.0_{\pm2.8}$ & 59.5 \\
RC-PE & $56.1_{\pm1.4}$ & $67.4_{\pm0.9}$ & $61.0_{\pm1.4}$ & 60.8 \\
Heatmap & $\mathbf{58.3}_{\pm0.0}$ & $66.9_{\pm0.7}$ & $61.0_{\pm1.4}$ & 61.7 \\
\midrule
\rowcolor{myblue}
GeoProp-Base & $57.1_{\pm0.8}$ & $\mathbf{75.6}_{\pm5.1}$ & $\mathbf{64.7}_{\pm0.9}$ & \textbf{64.8} \\
\bottomrule
\end{tabular}
\end{table}

\textbf{Small-object precision tasks.}
To support the claim in Sec.~\ref{sec:ablations} that GeoProp's gains are largest on precision-oriented tasks involving small objects, Table~\ref{tab:app_smallobj_precision} reports per-task results on four representative MetaWorld tasks under both Diffusion Policy backbones.
GeoProp improves \textit{Basketball}, \textit{Hand Insert}, \textit{Pick Place}, and \textit{Sweep Into} by \textbf{+24.0pp} on average with ResNet-18 and \textbf{+20.5pp} with ViT-Base, giving an overall mean gain of \textbf{+22.3pp}.
This indicates that explicit geometric grounding is especially helpful when the policy must reason about small visual targets, where vector-based proprioception often fails to localize the relevant interaction region.

\begin{table}[!htbp]
\centering
\caption{Per-task success rates (\%) on four representative small-object precision MetaWorld tasks (Diffusion Policy). $\Delta$ is the absolute gain of GeoProp over Vanilla.}
\label{tab:app_smallobj_precision}
\small
\setlength{\tabcolsep}{5.0pt}
\begin{tabular}{@{}l ccc ccc c@{}}
\toprule
\multirow{2}{*}{\textbf{Task}} & \multicolumn{3}{c}{\textbf{ResNet-18}} & \multicolumn{3}{c}{\textbf{ViT-Base}} & \multirow{2}{*}{\textbf{Avg $\Delta$}} \\
\cmidrule(lr){2-4}\cmidrule(lr){5-7}
& Vanilla & GeoProp & $\Delta$ & Vanilla & GeoProp & $\Delta$ & \\
\midrule
Basketball  & 8  & \textbf{38} & +30 & 16 & \textbf{38} & +22 & +26 \\
Hand Insert & 44 & \textbf{52} & +8  & 38 & \textbf{56} & +18 & +13 \\
Pick Place  & 12 & \textbf{36} & +24 & 2  & \textbf{30} & +28 & +26 \\
Sweep Into  & 64 & \textbf{98} & +34 & 76 & \textbf{90} & +14 & +24 \\
\midrule
\rowcolor{myblue}
\textbf{Mean} & 32.0 & \textbf{56.0} & \textbf{+24.0} & 33.0 & \textbf{53.5} & \textbf{+20.5} & \textbf{+22.3} \\
\bottomrule
\end{tabular}
\end{table}

\textbf{Expanded task suite.}
On the expanded 15-task RoboTwin setup (Table~\ref{tab:app_robotwin_15}), GeoProp improves $\pi_0$ from 66.4\% to 70.9\% (+4.5pp), comparable to the +4.0pp gain on the original 7-task setup, indicating that the benefit is not specific to the original task selection.

\textbf{Failure and low-gain cases.}
The expanded RoboTwin results also show that GeoProp is not uniformly beneficial on every task.
Gains are small on saturated tasks such as \emph{Adjust Bottle}, \emph{Click Alarmclock}, and \emph{Click Bell}, where Vanilla already reaches 100\% success.
GeoProp can also underperform on tasks such as \emph{Pick Diverse Bottles}, \emph{Place Shoe}, \emph{Scan Object}, and \emph{Stack Blocks Two}, suggesting that end-effector-only grounding is less reliable when the projected point is visually ambiguous, occluded, or not the dominant cue for task progress.
These cases are consistent with the limitation discussed in the main paper: projection provides a useful spatial prior only when the projected end-effector region remains informative for control.

\textbf{Parameter-matched control.}
A parameter-matched control that enlarges the proprioceptive encoder to match GeoProp's parameter count recovers only +0.5pp for Diffusion Policy and +0.1pp for $\pi_0$, indicating that the gains are driven primarily by geometric alignment rather than additional model capacity.

\begin{table}[!htbp]
\centering
\caption{Data-scaling results on the 7-task RoboTwin setup with $\pi_0$.}
\label{tab:app_data_scaling}
\small
\setlength{\tabcolsep}{6.0pt}
\begin{tabular}{@{}lccc@{}}
\toprule
\textbf{Demos / Task} & \textbf{Vanilla $\pi_0$} & \textbf{GeoProp} & \textbf{Gain} \\
\midrule
25  & 46.3 & 48.6 & +2.3 \\
50  & 64.0 & 68.0 & +4.0 \\
100 & 77.7 & 80.5 & +2.8 \\
\bottomrule
\end{tabular}
\end{table}

\begin{table}[!htbp]
\centering
\caption{Per-task success rates (\%) on the expanded 15-task RoboTwin setup with $\pi_0$ (50 demonstrations per task, 25 evaluation rollouts per task). $\Delta$ denotes the absolute gain of GeoProp over Vanilla $\pi_0$.}
\label{tab:app_robotwin_15}
\small
\setlength{\tabcolsep}{6.0pt}
\begin{tabular}{@{}lccc@{}}
\toprule
\textbf{Task} & \textbf{Vanilla $\pi_0$} & \textbf{GeoProp} & \textbf{$\Delta$} \\
\midrule
adjust\_bottle           & 100  & 100           & 0   \\
beat\_block\_hammer      & 68   & \textbf{92}   & +24 \\
click\_alarmclock        & 100  & 100           & 0   \\
click\_bell              & 100  & 100           & 0   \\
dump\_bin\_bigbin        & \textbf{92}   & 88   & -4  \\
move\_can\_pot           & 76   & \textbf{80}   & +4  \\
move\_playingcard\_away  & 76   & \textbf{80}   & +4  \\
open\_laptop             & 72   & \textbf{80}   & +8  \\
pick\_diverse\_bottles   & \textbf{36}   & 32   & -4  \\
pick\_dual\_bottles      & 68   & \textbf{76}   & +8  \\
place\_dual\_shoes       & 16   & \textbf{44}   & +28 \\
place\_shoe              & \textbf{64}   & 60   & -4  \\
scan\_object             & \textbf{28}   & 24   & -4  \\
stack\_blocks\_two       & \textbf{68}   & 64   & -4  \\
turn\_switch             & 32   & \textbf{44}   & +12 \\
\midrule
\rowcolor{myblue}
\textbf{Mean}            & 66.4 & \textbf{70.9} & \textbf{+4.5} \\
\bottomrule
\end{tabular}
\end{table}

\FloatBarrier
\subsection{Per-Task Success Rates on Simulation Benchmarks}
\label{app:per_task_sim}

To complement the aggregated numbers in Table~\ref{tab:dp_results}, we report per-task success rates for all 63 Diffusion Policy simulation tasks, covering RLBench (Table~\ref{tab:app_per_task_rlbench}), RoboTwin (Table~\ref{tab:app_per_task_robotwin_dp}), and MetaWorld-50 (Table~\ref{tab:app_per_task_metaworld}).
Each entry follows the protocol in Sec.~\ref{sec:sim_exp}; per-row bests are highlighted.
These tables make explicit that GeoProp's average-level gains are not concentrated on a small subset of tasks: across both ResNet-18 and ViT-Base backbones, GeoProp matches or improves over Vanilla on the large majority of individual tasks, and the largest gains are concentrated on visually demanding settings (e.g., \textit{Water Plants} and \textit{Put Rubbish In Bin} on RLBench; \textit{Move Can Pot}, \textit{Move Playingcard Away}, and \textit{Scan Object} on RoboTwin; \textit{Basketball}, \textit{Sweep Into}, \textit{Pick Place}, and \textit{Hand Insert} on MetaWorld), consistent with the small-object precision analysis in Sec.~\ref{sec:ablations}.

\begin{table}[!htbp]
\centering
\caption{Per-task success rates (\%) on RLBench with Diffusion Policy. Per-row bests are highlighted.}
\label{tab:app_per_task_rlbench}
\small
\setlength{\tabcolsep}{3.0pt}
\resizebox{\textwidth}{!}{%
\begin{tabular}{ll cccccc c}
\toprule
\multirow{2}{*}{\textbf{Backbone}} & \multirow{2}{*}{\textbf{Method}} &
\multicolumn{6}{c}{\textbf{RLBench}} & \multirow{2}{*}{\textbf{Mean}} \\
\cmidrule(lr){3-8}
& & Close Box & Put Rubbish In Bin & Close Laptop Lid & Water Plants & Unplug Charger & Toilet Seat Down & \\
\midrule
\multirow{3}{*}{ResNet-18}
& Vanilla    & \ddbf{100} & \dd{32}   & \dd{88}    & \dd{36}   & \dd{44}   & \dd{96}    & 66.0 \\
& No-Proprio & \ddbf{100} & \dd{44}   & \dd{92}    & \dd{28}   & \dd{48}   & \dd{96}    & 68.0 \\
& GeoProp    & \ddbf{100} & \ddbf{48} & \ddbf{100} & \ddbf{72} & \ddbf{56} & \ddbf{100} & \textbf{79.3} \\
\midrule
\multirow{3}{*}{ViT-Base}
& Vanilla    & \ddbf{100} & \dd{20}   & \dd{92}    & \dd{36}   & \dd{44}   & \ddbf{100} & 65.3 \\
& No-Proprio & \ddbf{100} & \dd{64}   & \dd{84}    & \dd{44}   & \dd{32}   & \ddbf{100} & 70.7 \\
& GeoProp    & \ddbf{100} & \ddbf{68} & \ddbf{100} & \ddbf{52} & \ddbf{56} & \ddbf{100} & \textbf{79.3} \\
\bottomrule
\end{tabular}%
}
\end{table}

\begin{table}[!htbp]
\centering
\caption{Per-task success rates (\%) on the RoboTwin 7-task suite with Diffusion Policy. Per-row bests are highlighted.}
\label{tab:app_per_task_robotwin_dp}
\small
\setlength{\tabcolsep}{3.0pt}
\resizebox{\textwidth}{!}{%
\begin{tabular}{ll ccccccc c}
\toprule
\multirow{2}{*}{\textbf{Backbone}} & \multirow{2}{*}{\textbf{Method}} &
\multicolumn{7}{c}{\textbf{RoboTwin}} & \multirow{2}{*}{\textbf{Mean}} \\
\cmidrule(lr){3-9}
& & Beat Block Hammer & Click Alarmclock & Move Can Pot & Move Playingcard Away & Place Shoe & Scan Object & Stack Blocks Two & \\
\midrule
\multirow{3}{*}{ResNet-18}
& Vanilla    & \ddbf{84} & \ddbf{84} & \dd{84}   & \dd{36}   & \dd{36}   & \dd{20}   & \ddbf{32} & 53.7 \\
& No-Proprio & \ddbf{84} & \ddbf{84} & \dd{72}   & \dd{48}   & \dd{24}   & \dd{40}   & \dd{24}   & 53.7 \\
& GeoProp    & \ddbf{84} & \dd{76}   & \ddbf{92} & \ddbf{72} & \ddbf{40} & \ddbf{44} & \ddbf{32} & \textbf{62.9} \\
\midrule
\multirow{3}{*}{ViT-Base}
& Vanilla    & \dd{68}   & \dd{72}   & \dd{68}   & \dd{24}   & \ddbf{32} & \dd{8}    & \ddbf{16} & 41.1 \\
& No-Proprio & \dd{64}   & \dd{60}   & \ddbf{76} & \dd{36}   & \dd{24}   & \dd{24}   & \ddbf{16} & 42.9 \\
& GeoProp    & \ddbf{88} & \ddbf{76} & \dd{68}   & \ddbf{52} & \ddbf{32} & \ddbf{36} & \dd{12}   & \textbf{52.0} \\
\bottomrule
\end{tabular}%
}
\end{table}

\begin{table}[!htbp]
\centering
\caption{Per-task success rates (\%) on the MetaWorld-50 suite with Diffusion Policy. Per-row bests are highlighted within each backbone.}
\label{tab:app_per_task_metaworld}
\small
\setlength{\tabcolsep}{4.0pt}
\resizebox{0.85\textwidth}{!}{%
\begin{tabular}{l|ccc|ccc}
\toprule
\multirow{2}{*}{\textbf{Task}} &
\multicolumn{3}{c|}{\textbf{ResNet-18}} &
\multicolumn{3}{c}{\textbf{ViT-Base}} \\
\cmidrule(lr){2-4}\cmidrule(lr){5-7}
& Vanilla & No-Proprio & GeoProp & Vanilla & No-Proprio & GeoProp \\
\midrule
Button Press              & \ddbf{100} & \ddbf{100} & \ddbf{100} & \ddbf{100} & \ddbf{100} & \ddbf{100} \\
Button Press Topdown      & \ddbf{100} & \ddbf{100} & \ddbf{100} & \ddbf{100} & \ddbf{100} & \ddbf{100} \\
Button Press Topdown Wall & \ddbf{100} & \ddbf{100} & \ddbf{100} & \ddbf{100} & \ddbf{100} & \ddbf{100} \\
Button Press Wall         & \ddbf{100} & \ddbf{100} & \ddbf{100} & \ddbf{100} & \ddbf{100} & \ddbf{100} \\
Coffee Button             & \ddbf{100} & \ddbf{100} & \ddbf{100} & \ddbf{100} & \ddbf{100} & \ddbf{100} \\
Dial Turn                 & \dd{96}    & \ddbf{98}  & \dd{88}    & \dd{58}    & \ddbf{62}  & \ddbf{62}  \\
Door Close                & \ddbf{100} & \ddbf{100} & \ddbf{100} & \ddbf{100} & \ddbf{100} & \ddbf{100} \\
Door Lock                 & \dd{64}    & \dd{68}    & \ddbf{74}  & \ddbf{68}  & \dd{70}    & \dd{54}    \\
Door Open                 & \ddbf{100} & \ddbf{100} & \ddbf{100} & \ddbf{100} & \ddbf{100} & \ddbf{100} \\
Door Unlock               & \ddbf{100} & \dd{98}    & \ddbf{100} & \dd{98}    & \dd{96}    & \ddbf{100} \\
Drawer Close              & \ddbf{100} & \ddbf{100} & \ddbf{100} & \ddbf{100} & \ddbf{100} & \ddbf{100} \\
Drawer Open               & \dd{74}    & \dd{78}    & \ddbf{82}  & \ddbf{76}  & \dd{72}    & \dd{74}    \\
Faucet Close              & \ddbf{100} & \ddbf{100} & \ddbf{100} & \ddbf{100} & \ddbf{100} & \ddbf{100} \\
Faucet Open               & \dd{84}    & \dd{82}    & \ddbf{96}  & \dd{78}    & \dd{78}    & \ddbf{94}  \\
Handle Press              & \ddbf{100} & \ddbf{100} & \ddbf{100} & \ddbf{100} & \ddbf{100} & \ddbf{100} \\
Handle Pull               & \dd{74}    & \dd{72}    & \ddbf{78}  & \dd{54}    & \dd{56}    & \ddbf{74}  \\
Handle Press Side         & \ddbf{100} & \ddbf{100} & \ddbf{100} & \dd{86}    & \dd{96}    & \ddbf{100} \\
Handle Pull Side          & \dd{70}    & \dd{72}    & \ddbf{74}  & \ddbf{70}  & \dd{56}    & \ddbf{74}  \\
Lever Pull                & \dd{6}     & \dd{8}     & \ddbf{10}  & \ddbf{14}  & \dd{8}     & \dd{8}     \\
Plate Slide               & \ddbf{100} & \ddbf{100} & \ddbf{100} & \ddbf{100} & \dd{98}    & \dd{98}    \\
Plate Slide Back          & \ddbf{100} & \ddbf{100} & \ddbf{100} & \ddbf{100} & \ddbf{100} & \ddbf{100} \\
Plate Slide Back Side     & \ddbf{100} & \ddbf{100} & \ddbf{100} & \ddbf{100} & \ddbf{100} & \ddbf{100} \\
Plate Slide Side          & \ddbf{98}  & \dd{96}    & \dd{94}    & \dd{98}    & \ddbf{100} & \ddbf{100} \\
Reach                     & \ddbf{12}  & \dd{8}     & \dd{8}     & \dd{8}     & \ddbf{10}  & \dd{8}     \\
Reach Wall                & \dd{34}    & \dd{38}    & \ddbf{42}  & \ddbf{46}  & \dd{40}    & \ddbf{46}  \\
Window Close              & \dd{92}    & \dd{84}    & \ddbf{100} & \dd{84}    & \dd{90}    & \ddbf{98}  \\
Window Open               & \ddbf{46}  & \dd{34}    & \ddbf{46}  & \dd{32}    & \dd{26}    & \ddbf{52}  \\
Peg Unplug Side           & \ddbf{40}  & \dd{28}    & \dd{30}    & \dd{22}    & \ddbf{40}  & \ddbf{40}  \\
Basketball                & \dd{8}     & \dd{6}     & \ddbf{38}  & \dd{16}    & \dd{14}    & \ddbf{38}  \\
Bin Picking               & \dd{66}    & \dd{70}    & \ddbf{78}  & \dd{66}    & \ddbf{72}  & \dd{64}    \\
Box Close                 & \ddbf{48}  & \dd{36}    & \dd{40}    & \ddbf{54}  & \ddbf{54}  & \ddbf{54}  \\
Coffee Pull               & \dd{74}    & \dd{66}    & \ddbf{84}  & \dd{70}    & \dd{74}    & \ddbf{86}  \\
Coffee Push               & \dd{82}    & \dd{84}    & \ddbf{90}  & \dd{52}    & \dd{46}    & \ddbf{74}  \\
Hammer                    & \dd{94}    & \dd{94}    & \ddbf{96}  & \ddbf{92}  & \dd{90}    & \ddbf{92}  \\
Peg Insert Side           & \dd{26}    & \dd{36}    & \ddbf{38}  & \dd{8}     & \dd{20}    & \ddbf{24}  \\
Push Wall                 & \dd{44}    & \dd{44}    & \ddbf{88}  & \dd{48}    & \dd{64}    & \ddbf{92}  \\
Soccer                    & \dd{14}    & \dd{12}    & \ddbf{16}  & \ddbf{12}  & \ddbf{12}  & \dd{6}     \\
Sweep                     & \dd{20}    & \dd{32}    & \ddbf{76}  & \dd{20}    & \dd{30}    & \ddbf{68}  \\
Sweep Into                & \dd{64}    & \dd{82}    & \ddbf{98}  & \dd{76}    & \dd{80}    & \ddbf{90}  \\
Assembly                  & \dd{96}  & \dd{92}    & \ddbf{98}  & \dd{50}    & \dd{48}    & \ddbf{82}  \\
Hand Insert               & \dd{44}    & \dd{40}    & \ddbf{52}  & \dd{38}    & \dd{24}    & \ddbf{56}  \\
Pick Out Of Hole          & \dd{42}    & \ddbf{56}  & \dd{50}    & \dd{32}    & \dd{30}    & \ddbf{42}  \\
Pick Place                & \dd{12}    & \dd{16}    & \ddbf{36}  & \dd{2}     & \dd{6}     & \ddbf{30}  \\
Push                      & \dd{34}    & \dd{50}    & \ddbf{66}  & \dd{34}    & \dd{30}    & \ddbf{58}  \\
Push Back                 & \dd{48}    & \ddbf{52}  & \ddbf{52}  & \ddbf{48}  & \dd{46}    & \ddbf{48}  \\
Pick Place Wall           & \dd{40}    & \dd{68}    & \ddbf{86}  & \dd{70}    & \dd{36}    & \ddbf{76}  \\
Stick Pull                & \dd{74}    & \dd{80}    & \ddbf{86}  & \ddbf{74}  & \dd{58}    & \dd{72}    \\
Stick Push                & \ddbf{100} & \ddbf{100} & \ddbf{100} & \ddbf{100} & \ddbf{100} & \ddbf{100} \\
Shelf Place               & \dd{46}    & \dd{46}    & \ddbf{60}  & \dd{14}    & \dd{14}    & \ddbf{30}  \\
Disassemble               & \dd{72}    & \dd{78}    & \ddbf{80}  & \dd{60}    & \dd{56}    & \ddbf{68}  \\
\midrule
\rowcolor{myblue}
\textbf{Mean (50)}        & 68.8 & 70.1 & \textbf{76.6} & 64.6 & 64.0 & \textbf{72.6} \\
\bottomrule
\end{tabular}%
}
\end{table}

\FloatBarrier
\subsection{Qualitative Grounding Visualization}
\label{app:qual_grounding}

\begin{figure}[!htbp]
    \centering
    \includegraphics[width=0.68\textwidth]{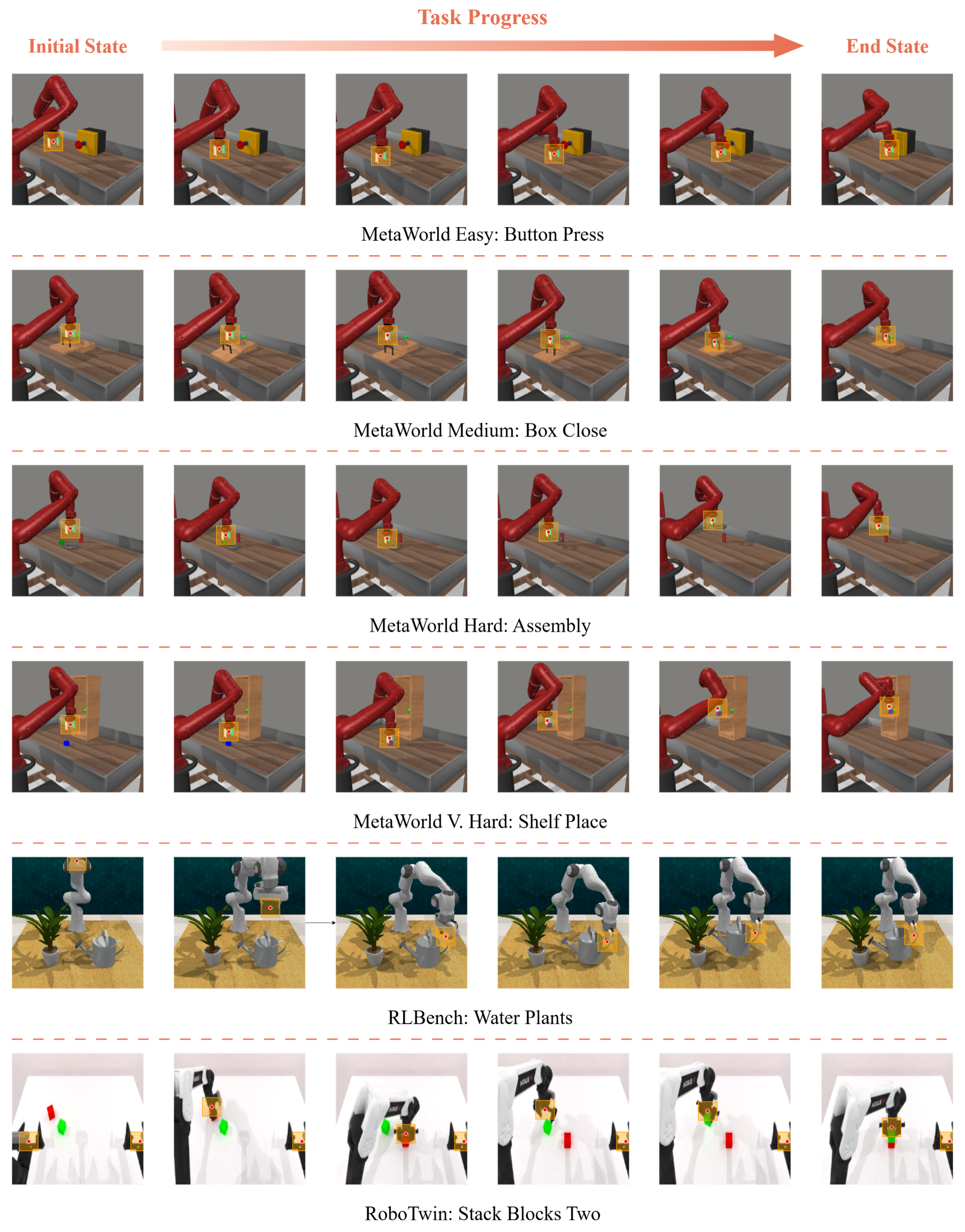}
    \caption{\textbf{Simulation grounding visualization.}
    Rows show representative rollouts from simulation tasks.
    Red dots indicate projected 2D end-effector positions, and orange boxes show the corresponding image-space regions associated with GeoProp's feature-grid grounding.}
    \label{fig:sim_grounding_vis}
\end{figure}

We provide qualitative visualizations to illustrate how GeoProp grounds proprioceptive state in visual feature space.
Fig.~\ref{fig:sim_grounding_vis} shows representative simulation rollouts with projected end-effector locations and their corresponding image-space grounding regions.
The red dot denotes the 2D projection of the 3D end-effector position, and the orange box visualizes the image-space region associated with the feature-grid cell used by GeoProp.
These examples show that the projected robot state remains spatially aligned with task-relevant interaction regions across different simulated embodiments and manipulation tasks.

\begin{figure}[!htbp]
    \centering
    \includegraphics[width=\textwidth]{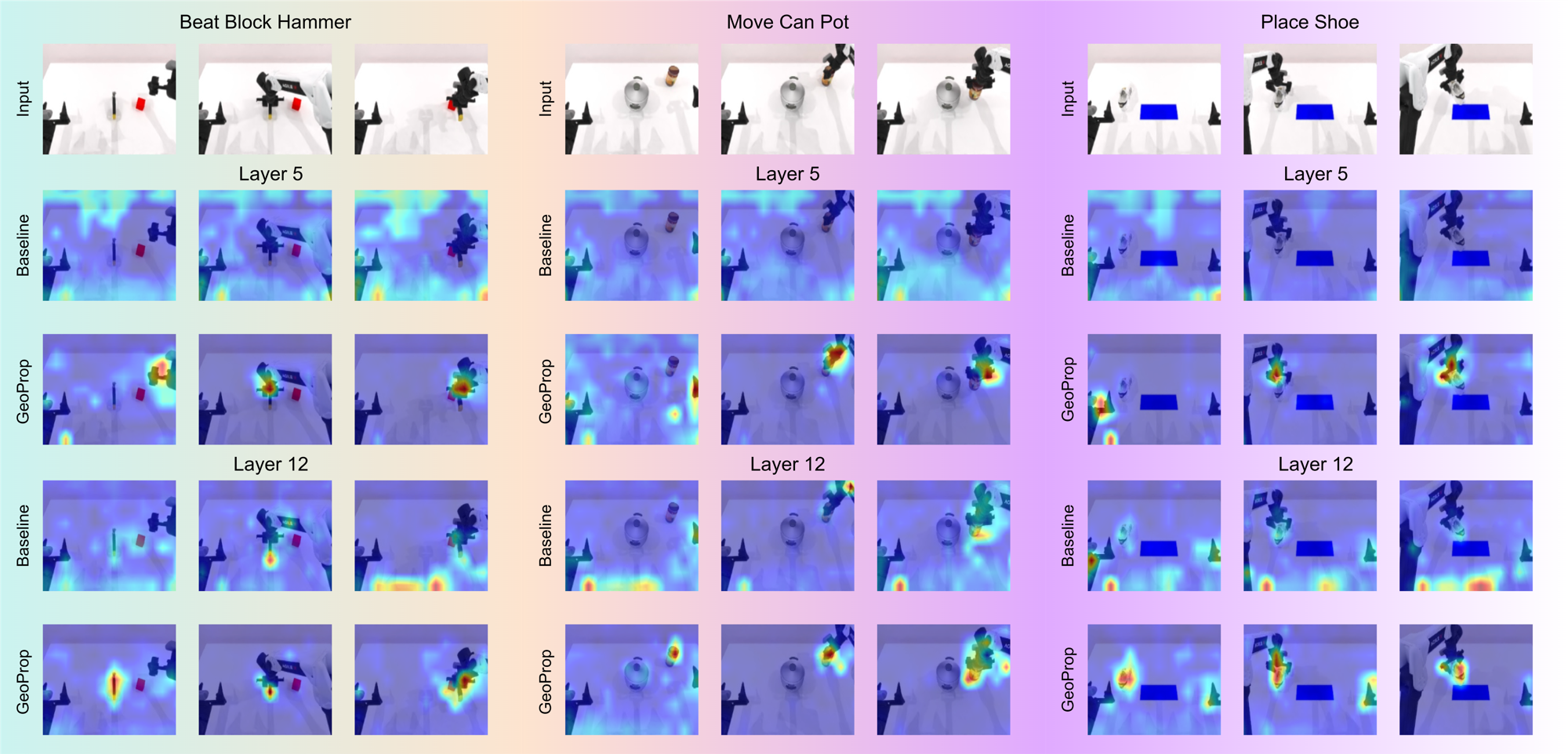}
    \caption{\textbf{Additional attention-map visualizations.}
    We compare Vanilla proprioception fusion and GeoProp across multiple RoboTwin tasks and transformer layers.
    GeoProp produces more localized activations around task-relevant end-effector/object regions, while Vanilla fusion often attends diffusely to background regions.}
    \label{fig:attention_maps_full}
\end{figure}

Fig.~\ref{fig:attention_maps_full} further compares attention maps between Vanilla proprioception fusion and GeoProp across multiple RoboTwin tasks and transformer layers.
Vanilla fusion often produces diffuse activations over background regions, whereas GeoProp yields more localized responses around the end-effector and task-relevant objects.
This supports our central hypothesis that explicit geometric grounding helps the policy associate proprioceptive state with co-located visual evidence, rather than treating state as a disjoint global vector.

\FloatBarrier
\subsection{Calibration Robustness}
\label{app:calibration_robustness}

GeoProp relies on camera--robot calibration to project 3D end-effector states into the image plane.
To characterize this assumption, we evaluate GeoProp under synthetic test-time perturbations to the camera extrinsics.
All policies are trained under the same setting, and perturbations are applied only during evaluation.
We compare GeoProp with Vanilla proprioception fusion and No-Proprio using the same 15-task MetaWorld protocol as in the ablation study.

\begin{figure}[!htbp]
    \centering
    \begin{subfigure}[t]{0.49\textwidth}
        \centering
        \includegraphics[width=\linewidth]{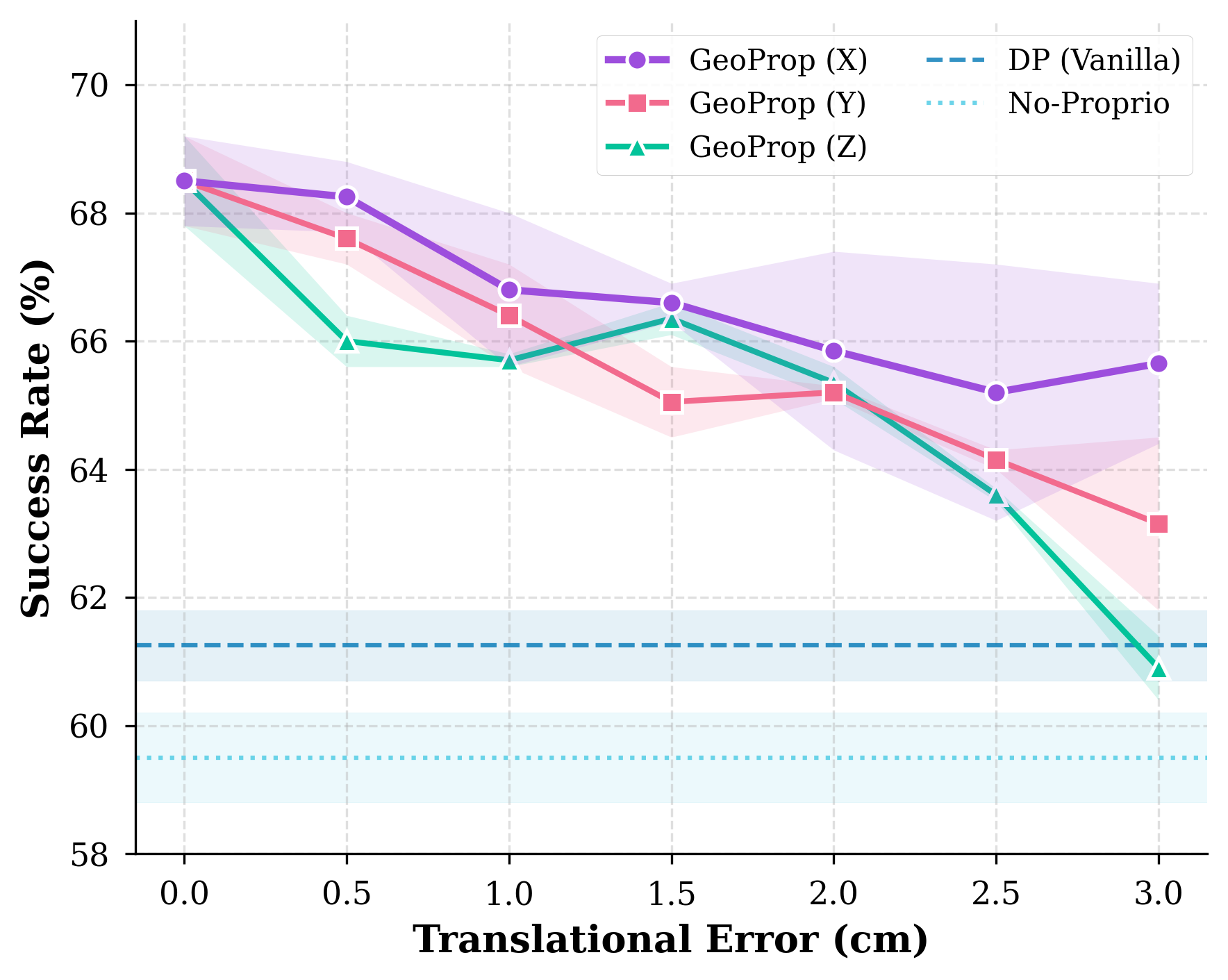}
        \caption{Translation drift.}
        \label{fig:robustness_translation_app}
    \end{subfigure}
    \hfill
    \begin{subfigure}[t]{0.49\textwidth}
        \centering
        \includegraphics[width=\linewidth]{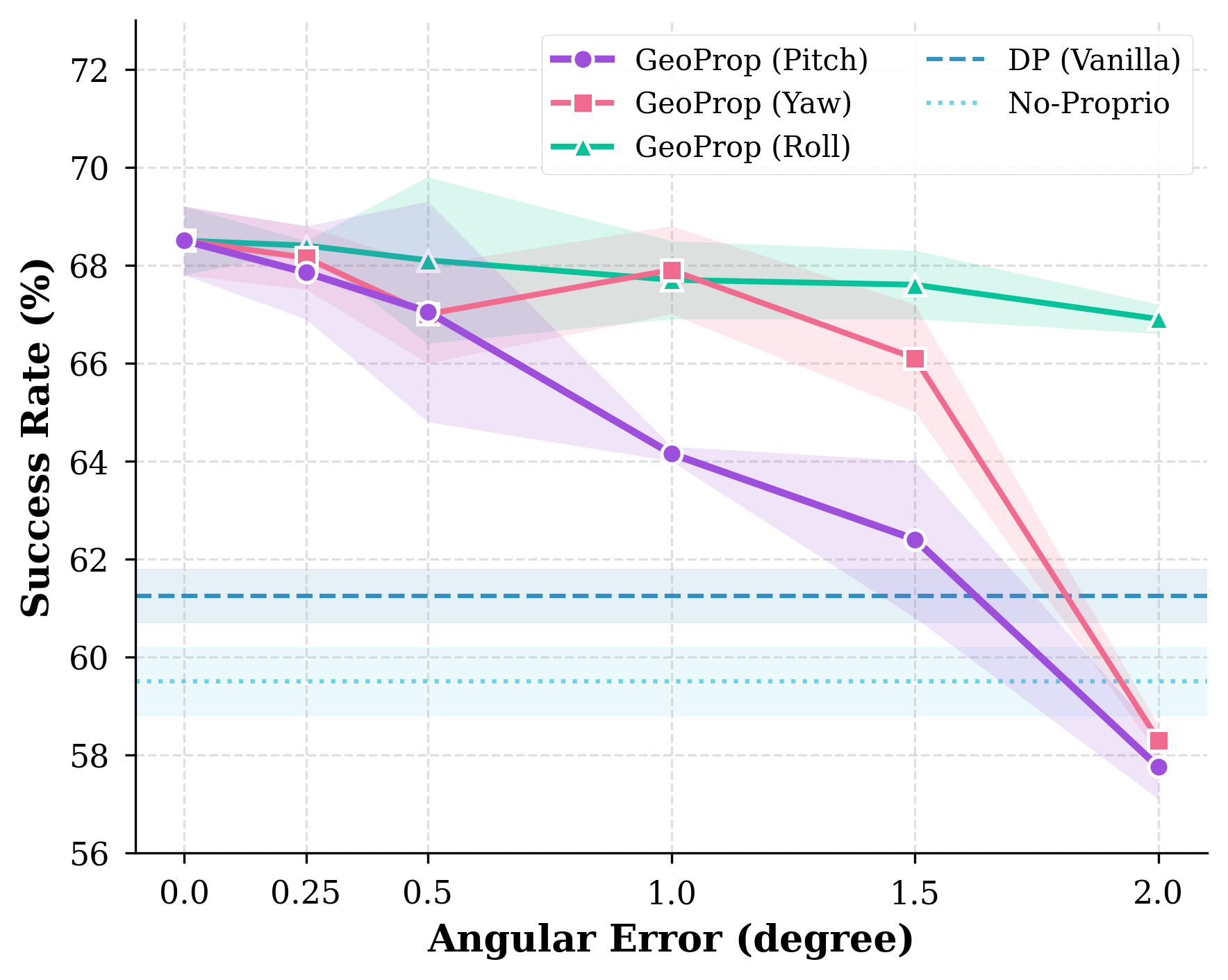}
        \caption{Rotation drift.}
        \label{fig:robustness_rotation_app}
    \end{subfigure}
    \caption{\textbf{Calibration robustness under synthetic extrinsic drift.}
    GeoProp remains above Vanilla and No-Proprio under mild-to-moderate perturbations, while performance gradually decreases as projection error grows.}
    \label{fig:calibration_robustness_app}
\end{figure}

For translation drift, we perturb the camera--robot translation along each Cartesian axis independently.
For rotation drift, we perturb the camera orientation around roll, pitch, and yaw.
The dashed horizontal lines denote the mean performance of Vanilla and No-Proprio baselines, which are unaffected by projection noise.
As shown in Fig.~\ref{fig:calibration_robustness_app}, GeoProp degrades gradually under increasing calibration error.
It remains above both baselines under mild-to-moderate translation drift up to 2.5\,cm and rotation drift up to $1.5^\circ$.
The degradation is axis-dependent: translation errors along the depth-related axis and rotation errors around pitch/yaw cause larger drops, while roll drift is substantially less harmful.
These results show that GeoProp benefits from accurate geometric grounding, but remains usable under moderate calibration noise.

\begin{figure}[!htbp]
    \centering
    \includegraphics[width=\textwidth]{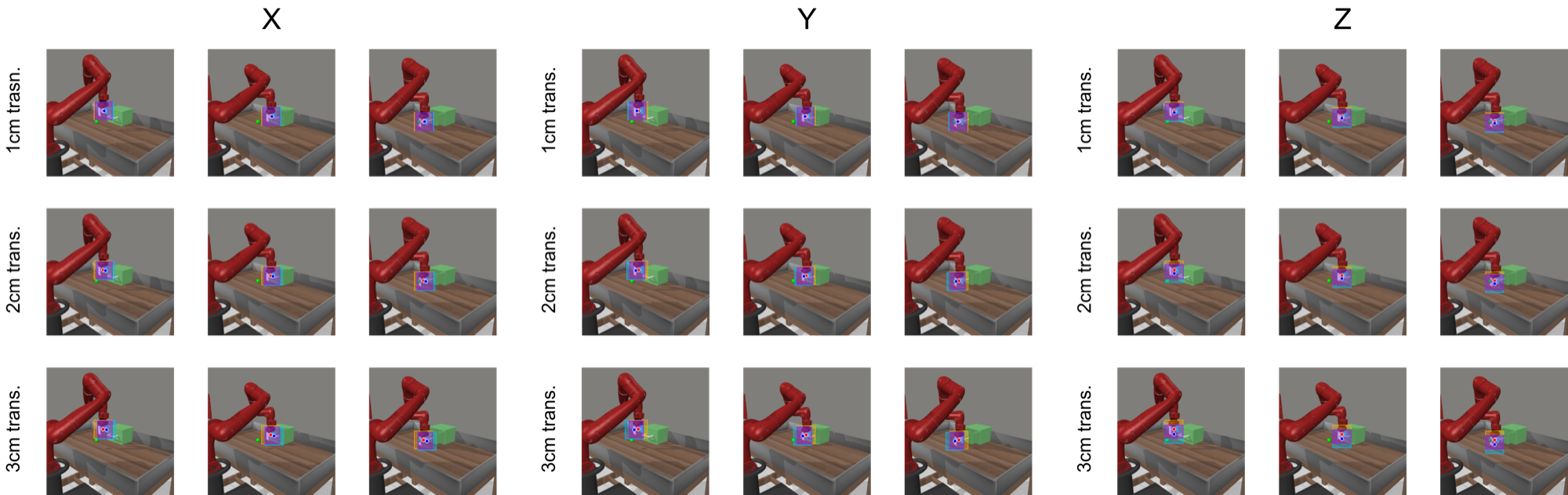}
    \caption{\textbf{Qualitative visualization under translation drift.}
    Red and blue dots show unperturbed and perturbed end-effector projections, respectively.
    Orange and cyan boxes show the corresponding grounding regions, with overlap highlighted in purple.}
    \label{fig:calib_error_trans_vis}
\end{figure}

\begin{figure}[!htbp]
    \centering
    \includegraphics[width=\textwidth]{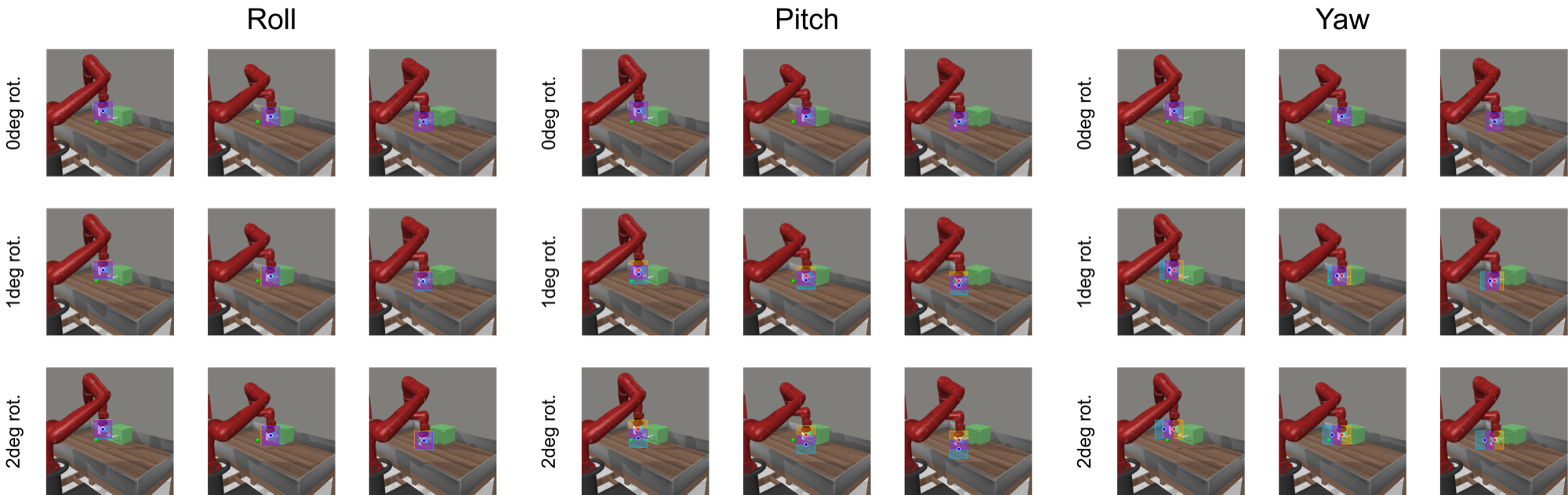}
    \caption{\textbf{Qualitative visualization under rotation drift.}
    Red and blue dots show unperturbed and perturbed end-effector projections, respectively.
    Orange and cyan boxes show the corresponding grounding regions, with overlap highlighted in purple.}
    \label{fig:calib_error_rot_vis}
\end{figure}

To visualize how calibration drift affects GeoProp's spatial grounding, we further show projected locations and sampled grounding regions under perturbed extrinsics.
In Figs.~\ref{fig:calib_error_trans_vis} and~\ref{fig:calib_error_rot_vis}, the red dot indicates the end-effector projection under the original extrinsics, while the blue dot indicates the projection under perturbed extrinsics.
The orange box denotes the grounding region from the original projection, the cyan box denotes the grounding region from the perturbed projection, and the purple region shows their overlap.
As calibration error increases, the perturbed grounding region gradually shifts away from the original one, explaining the performance degradation observed in Fig.~\ref{fig:calibration_robustness_app}.

\FloatBarrier
\section{Real-World Experiments Details}
\label{app:real_world_details}

\subsection{Qualitative Execution Sequences}
\label{sec:real_world_qualitative}

\begin{figure}[!htbp]
    \centering
    \includegraphics[width=\textwidth]{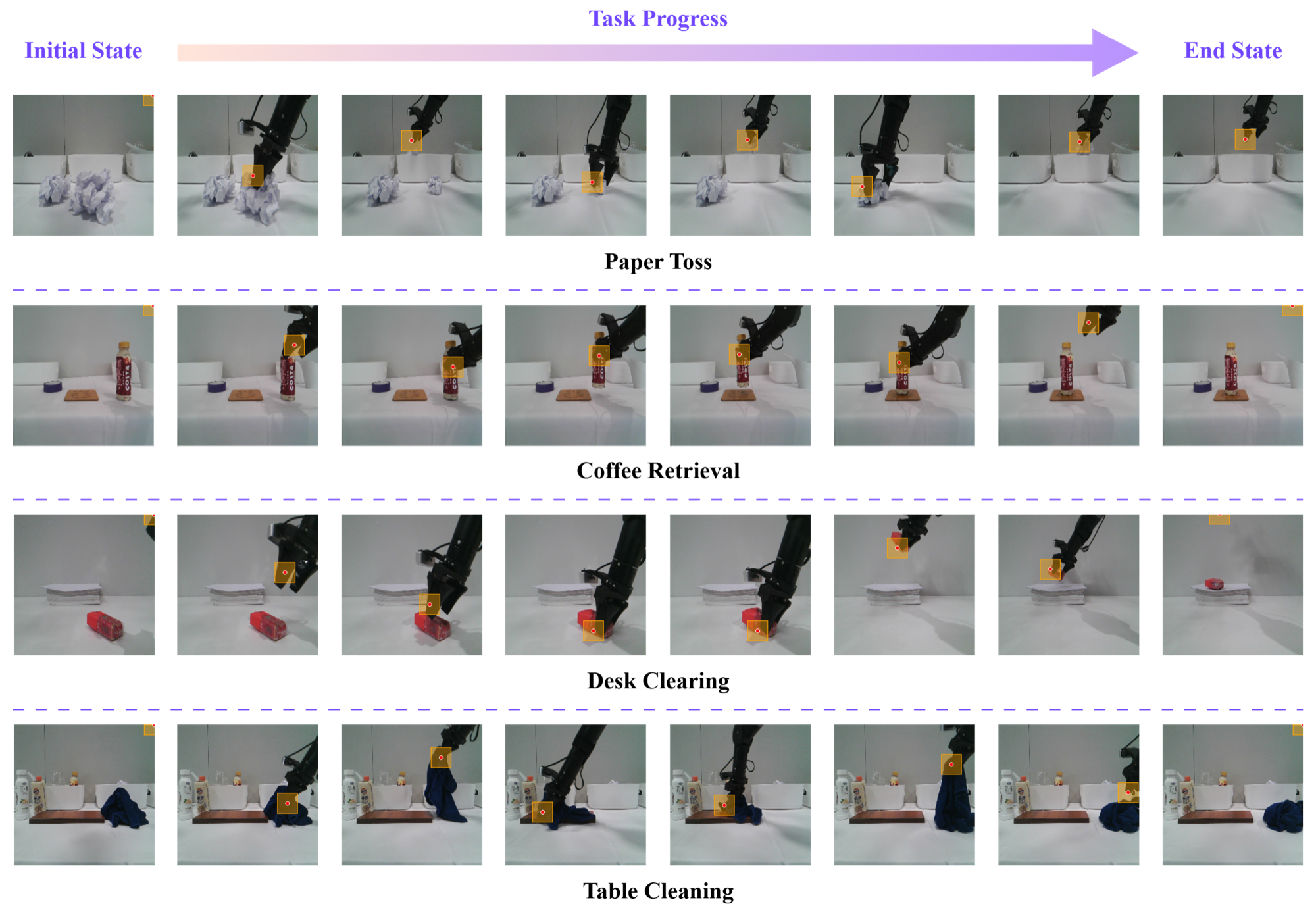}
    \caption{\textbf{Real-world execution sequences on Mobile ALOHA.}
    Rows show task progress from the initial state to the final state for Paper Toss, Coffee Retrieval, Desk Clearing, and Table Cleaning.
    Red dots indicate projected end-effector positions, and orange boxes show the corresponding grounded visual regions used by GeoProp.}
    \label{fig:real_world_execution_app}
\end{figure}

Fig.~\ref{fig:real_world_execution_app} shows representative real-world execution sequences on the Mobile ALOHA platform.
Each row corresponds to one household manipulation task and visualizes task progress from the initial state to the final state.
The red dot denotes the projected 2D end-effector position, and the orange box indicates the image-space region corresponding to GeoProp's feature-grid grounding.
These examples illustrate that GeoProp can maintain spatial correspondence between proprioceptive state and task-relevant visual evidence across grasping, object clearing, and wiping behaviors.

\end{document}